\documentclass[]{spie}  

\usepackage{amsmath,amsfonts,amssymb}
\usepackage{graphicx}
\usepackage[colorlinks=true, allcolors=blue]{hyperref}
\usepackage{subcaption}
\usepackage{caption}
\usepackage{booktabs}
\usepackage{multirow}
\usepackage{makecell} 
\usepackage{arydshln} 
\usepackage{adjustbox}
\usepackage{amssymb}
\usepackage{algorithm}
\usepackage{algpseudocode}

\title{UGC Quality Assessment: Exploring the Impact of Saliency in Deep Feature-Based Quality Assessment}

\author{Xinyi Wang}
\author{Angeliki Katsenou}
\author{David Bull}
\affil{Visual Information Laboratory, University of Bristol, BS1 5DD, Bristol, UK}

\authorinfo{Further author information: Send correspondence to Xinyi Wang\\Xinyi Wang, E-mail: xinyi.wang@bristol.ac.uk\\  }

\begin{document} 
\maketitle

\begin{abstract}
The volume of User Generated Content (UGC) has increased in recent years. The challenge with this type of content is assessing its quality. So far, the state-of-the-art metrics are not exhibiting a very high correlation with perceptual quality. In this paper, we explore state-of-the-art metrics that extract/combine natural scene statistics and deep neural network features. We experiment with these by introducing saliency maps to improve perceptibility. We train and test our models using public datasets, namely, YouTube-UGC\cite{wang2019youtube} and KoNViD-1k\cite{hosu2017konstanz}. Preliminary results indicate that high correlations are achieved by using only deep features while adding saliency is not always boosting the performance. Our results and code will be made publicly available to serve as a benchmark for the research community and can be found on our project page \footnote{\url{https://github.com/xinyiW915/SPIE-2023-Supplementary}}.

\end{abstract}


\section{INTRODUCTION}
\label{sec: intro}  
User-generated videos have become an increasingly popular form of streaming on various platforms such as Instagram, TikTok, and Facebook, uploaded by creators and shared by users. The exponential growth of user-generated content (UGC) has propelled research in this area to the forefront~\cite{che2015survey}. However, it is crucial to note that a substantial portion of UGC videos lack professional production quality, unlike traditional professionally produced videos. In the UGC use case, videos are typically first encoded at the user's device and then transcoded by the platforms for storage and distribution. As a result of the repeated lossy compression, UGC videos suffer from the presence of noticeable artefacts such as blocking, blurring, ringing, etc.~\cite{Adsumilli2019, wang2019youtube}. 

A major challenge for the video distribution platform of such content is the quality assessment as the original uncompressed version that could serve as a reference is not available. Consequently, traditional metrics such as Peak Signal-to-Noise Ratio (PSNR), Structural Similarity Index (SSIM)\cite{wang2004image}, and Video Multimethod Assessment Fusion (VMAF)\cite{li2016toward} cannot be employed for UGC content. Therefore, the design of Reduced-Reference (RR) or Non-Reference (NR) metrics becomes imperative in assessing this type of content. These metrics primarily focus on analyzing pixel changes resulting from video distortions and employ algorithms, such as machine learning, to establish a correlation between distortion characteristics and subjective quality scores. These metrics enable the development of a predictive model for assessing video distortion degrees. Generally, NR metrics are more practical as they do not require any side information to be transmitted. The application of blind video quality metrics holds substantial value for video service providers such as YouTube and TikTok, enabling them to ensure a high-quality viewing experience for their end users. Furthermore, ideally, the Blind Video Quality Assessment (BVQA) model for UGC videos should exhibit a strong correlation with perceptual quality.

Among the traditional metrics, the use of Natural Scene Statistics (NSS) is the basis of many BVQA models, such as the Natural Image Quality Evaluator (NIQE)\cite{mittal2012making} and Blind Image Spatial Quality Evaluator (BRISQUE)\cite{mittal2011blind}. These metrics primarily rely on natural scene statistics in the spatio-temporal and/or frequency domain. Additionally, other metrics have been developed around the quantification of compression artifacts such as Noise\cite{chen2016perceptual}, Banding\cite{wang2016perceptual}, and SLEEQ\cite{ghadiyaram2017no} that are used for the characterisation of the YouTube UGC dataset\cite{wang2019youtube}.

Over the past decade, advancements in machine learning/deep learning research have led to the development of various learning-based methods. For instance, Le Kang~\cite{kang2014convolutional} employed a Convolutional Neural Network (CNN) with image patches as input, eliminating the need for hand-crafted features used in most previous approaches. Similarly, Li et al.~\cite{li2019quality} proposed a BVQA model by integrating both content-dependency and temporal-memory effects into a deep neural network (VSFA), which extracts features from a pre-trained image classification neural network to achieve content dependence. More recently, Google’s UGC-VQA\cite{tu2021ugc} introduced a feature selection strategy and established a fusion-based BVQA model (VIDEVAL), effectively balancing performance and efficiency trade-offs. Another metric from the same team, RAPIQUE~\cite{tu2021rapique}, combines natural scene statistics with semantic-aware deep convolution functions, resulting in reduced computational costs and runtime. Given that these metrics rely only on visual information and do not demonstrate a very high correlation with perceptual quality, there is room for further improvement. 

In an effort to improve the correlation to the perceptual quality of some of the best performing learning-based metrics, we are including amongst the fused features saliency maps that are generally aligned with fixation areas. The first step of this study is therefore to perform a benchmarking exercise on the current state-of-the-art BVQA models available for UGC videos and to explore the correlation between perceptual quality and BVQA models. Further, we experiment with the considered BVQA models by introducing saliency maps as additional features to test whether this would improve perceptual alignment. To evaluate the effectiveness of these hybrid models, we conduct training and testing using publicly available datasets, namely YouTube-UGC and KoNViD-1k. Preliminary results show that improved Spearman correlation (SRCC) with subjective scores can be achieved by using deep spatial features alone, while the inclusion of saliency does not consistently enhance performance. 

The structure of the rest of this paper is as follows: Section~\ref{sec: bg} presents the background and related research on video features, video quality evaluation algorithms, and visual saliency maps. Following, Section~\ref{sec: method} outlines the study's overall methodology and details the experimental tasks designed. Section~\ref{sec: res} reports on the experimental findings. The paper concludes with Section~\ref{sec: con} which highlights the contributions of this study and explores future research directions.

\section{Background}
\label{sec: bg}
Research around BVQA metrics is important as it can assess the effectiveness of video compression and streaming methodologies, as well as drive optimisation across the video streaming pipeline. In the following paragraphs, we provide a brief overview of the state-of-the-art BVQA metrics, as well as methods to extract saliency maps.

\textbf{Traditional BVQA Metrics.} 
Many BVQA models have been successfully developed within traditional metrics based on NSS, such as NIQE by Mittal et al.~\cite{mittal2012making}, which assesses image quality without prior knowledge of expected distortion or human opinion. It takes into account colour, frequency, temporal and spatial correlations, as well as statistical distributions. Similarly, Zhang et al.~\cite{7094273} proposed the Integrated Local NIQE (IL-NIQE) algorithm by combining structural statistical features, multi-scale orientation and frequency statistics, and colour statistical features. Moorthy et al.~\cite{moorthy2011blind} introduced PIQE, and Mittal et al.~\cite{mittal2011blind} developed BRISQUE based on spatial domain NSS. The difference between these two metrics compared to NIQE is the features they use to assess image quality. NIQE analyses the statistical features of an image, while PIQE simulates the human visual system, prioritising perceptual quality. On the other hand, BRISQUE focuses on spatial domain features and analyses statistical features within local image blocks. Other well-known BVQA tools include V-BLINDS~\cite{6705673} and V-CORNIA\cite{7025098}. V-CORINIA represented the features of image blocks by learning codebook, whereas V-BLINDS used Discrete Cosine Transform (DCT) statistics to mark image integrity. Both methods exhibit high sensitivity to specific types of distortion and are reliant on data quality. However, BVQA is still a complex task that includes different application scenarios and features. Existing methods encounter limitations such as computational complexity and generalisation capabilities. By combining machine learning and deep learning techniques, to improve the accuracy and generalisation ability to adapt to the needs of different application scenarios.

\textbf{Learning-based BVQA Metrics.}
A high number of quality metrics in the literature use Machine Learning (ML). These ML-based metrics focus on designing effective visual feature representations by learning a mapping model from visual features to visual quality through classical machine learning methods\cite{6272356,8412573,utke2022ndnetgaming} such as Support Vector Regression (SVR). However, these methods can require large amounts of annotated data and computational resources and may perform inconsistently in new scenarios. Deep neural networks, particularly CNNs, have gained significant attention in BVQA research. The primary advantage of deep neural networks is their capability to acquire abstract visual features. It is able to learn various representations of the video datasets without the need to manually design feature extractors, making the model more flexible and adaptable. 

Deep neural networks can learn visual features directly from distorted images or videos. The temporal aspect of videos adds complexity and requires usually more complex networks to address both spatial and temporal aspects. For example, the method proposed by Le Kang~\cite{kang2014convolutional} is more suitable for video analysis, as the 3D convolution effectively preserves temporal information. Liu et al.~\cite{liu2018end} proposed the video multitasking end-to-end optimization neural network V-MEON, which extends the multitasking end-to-end optimization neural network (MEON)~\cite{ma2017end} to predict quality scores by optimizing the feature extraction part through the regression. Recent innovations include VSFA~\cite{li2019quality}, which extracts features from a pre-trained image classification neural network to incorporate content dependency, and Two-Level Video Quality Model (TL-VQM)~\cite{korhonen2019two} employs a two-level feature extractor to efficiently compute distortion-related features. In addition, Long Short-Term Memory Networks (LSTM) have also been introduced to capture the long-term time dependence in video quality prediction~\cite{varga2019no}. Feng et al.~\cite{feng2022rankdvqa} introduced a BVQA method based on two-stage training using quality ranking to train a new transformer network architecture that avoids the use of manually annotated labels. The large-scale training database also optimises deep networks with high quality evaluation capability.

For the UGC video quality assessment study, Google adopted a feature selection strategy to balance performance and efficiency, selecting 60 out of 763 commonly used video statistical features to build a fusion-based BVQA model (VIDEVAL)~\cite{tu2021ugc} for benchmarking. However, its scalability for high-resolution and high-frame-rate videos remains a challenge. Another noteworthy metric is RAPIQUE~\cite{tu2021rapique}, which combines NSS with deep features extracted from a CNN to provide state-of-the-art performance on large-scale UGC datasets while reducing computational cost and runtime.  The idea of adding semantic information in the video quality metrics has been explored by a recently released model, UVQ\cite{wang2021rich}. Along with this model, an enhanced dataset was introduced, namely the YT-UGC$^+$ dataset, that is augmenting the existing YouTube-UGC dataset with content labels, compressed videos, and different Mean Opinion Scores (MOS). Their DNN-based framework analyses content, technical quality, and compression level to provide quality scores and human-friendly indicators, bridging low-level video signals to human perceptual quality. Although the pre-trained models on the YT-UGC$^+$ dataset have been released, the source code and the YT-UGC$^+$ dataset have not been shared publicly.

\textbf{Saliency Features.}
Saliency detection based on image processing techniques and computer vision algorithms can identify salient and noticeable regions in images. Research on saliency target detection began to receive widespread attention~\cite{liu2010learning} after the publication of literature by Itti et al.\cite{itti1998model} in 1998. While saliency has been extensively investigated in image quality assessment, its application in video quality assessment is relatively limited due to the high cost of capturing human eye movement information. Nevertheless, Jiang et al.~\cite{jiang2018deepvs} and other studies developed a deep learning-based video saliency detection method that combines static and dynamic structures to predict intra- and inter-frame saliency. Deep learning models have demonstrated superior performance compared to traditional models in predicting saliency~\cite{9191203}. Generating saliency maps provides valuable insights into the internal mechanisms and decisions of deep neural networks. Wang et al.~\cite{wang2020score} introduced Score-CAM, a weighted visual interpretation method for convolutional neural networks, which generates smoother saliency maps compared to gradient-based methods.

\section{Methodology}
\label{sec: method}
This paper comprehensively studies state-of-the-art BVQA metrics, including methods based on natural scene statistics and deep neural network features, such as RAPIQUE\cite{tu2021rapique} and VSFA\cite{li2019quality}. We chose these two state-of-the-art BVQA metrics as they are both prominent and effective in terms of accuracy and performance, with RAIQUE achieving high accuracy in faster runtimes. They focus on NSS features and deep neural network features. On this basis, we explored how feature fusion can be performed to enable comparisons with different feature combinations. Then, we introduced saliency maps generated using Score-CAM\cite{wang2020score} to fuse saliency features into the BVQA metrics to explore whether the perceptual relevance of these assessment methods is enhanced. To gain more insight into the contribution of saliency, deep features, and natural scene statistical features, we conducted an ablation study to understand and quantify their effects on blind quality assessment. These experiments allow us to understand and quantify the effects of different features on BVQA in depth, providing more comprehensive guidance and insights for subsequent research. 

\subsection{Baseline Models}
\label{subsec: baseline}
\textbf{RAPIQUE}\cite{tu2021rapique}. It uses a two-branch framework that combines quality-aware low-level features (NSS features) and semantic-aware high-level features (CNN features). The NSS features considered are mostly based on simple yet effective low-order bandpass statistics, such as the Mean Subtraction and Contrast Normalisation (MSCN), the Asymmetric Generalised Gaussian Distribution (AGGD) and Generalised Gaussian Distribution (GGD) to extract the features in the spatial domain across different directions. In the temporal domain, Haar wavelet filters are utilized to extract temporal features from the video. The deep learning feature extractor employed a pre-trained ResNet-50 as the semantic feature extractor, which is adapted to ($224\times224$) input dimensions through downsampling. Sparse frame sampling is used in order to reduce the computation time. In the spatial domain, spatial NSS features were extracted (2 fps) resulting in 680 dimensional features. In the temporal domain, temporal NSS features were extracted (8 fps) to obtain 476 dimensional features. The CNN features were extracted from one frame every second. Overall, the RAPIQUE feature extraction process includes spatial NSS features, temporal NSS features, and deep learning features that generate 3884-dimensional feature vectors for each video chunk after average pooling. 

\noindent
\textbf{VSFA}\cite{li2019quality}. It focuses on two parts: content-aware feature extraction and temporal-memory effects modelling. The former consists of feeding all video frames into a pre-trained CNN network (ResNet-50) and obtaining content-aware features (4096 dimensions) by spatial pooling of the corresponding feature maps. The latter inputs the video features into the GRU layer to capture the long-time dependencies between frames to get the integrated features to predict the subjective scores. In this paper, we focus more on the content-aware feature extractor in the VSFA model, which captures the content characteristics by extracting deep features from pre-trained CNNs. 

In addition to these state-of-the-art metrics, we explored the impact of deep learning features as BVQA metrics. We extracted deep features using only the content-aware feature extractor in the VSFA model (omitting the temporal dependency features), which will be henceforth denoted as VSFA\_CNN, and predicted quality scores using the same SVR regression classifier, as in the RAPIQUE model. 
\begin{figure}[ht]
    \centering

    \begin{minipage}{0.25\textwidth}
        \centering
        \includegraphics[height=2cm]{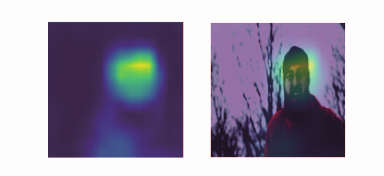}
        \includegraphics[height=2cm]{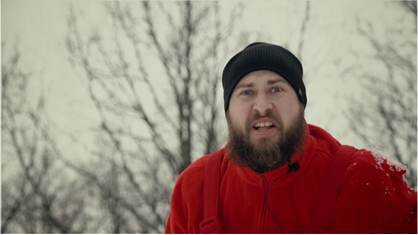}
    \captionsetup{font=footnotesize}
        \subcaption{Sampled Frame 1.}
        \label{fig: sf1}
    \end{minipage}%
    \begin{minipage}{0.25\textwidth}
        \centering
        \includegraphics[height=2cm]{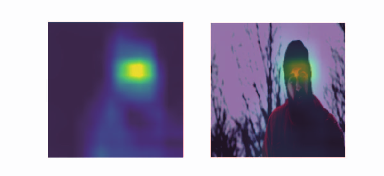}
        \includegraphics[height=2cm]{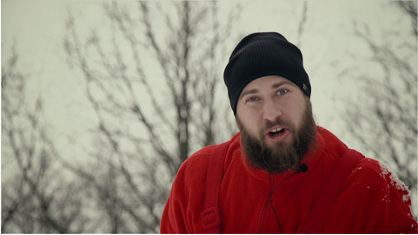}
    \captionsetup{font=footnotesize}
        \subcaption{Sampled Frame 2.}
        \label{fig: sf2}
    \end{minipage}%
    \begin{minipage}{0.25\textwidth}
        \centering
        \includegraphics[height=2cm]{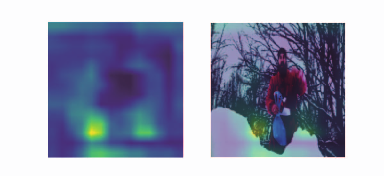}
        \includegraphics[height=2cm]{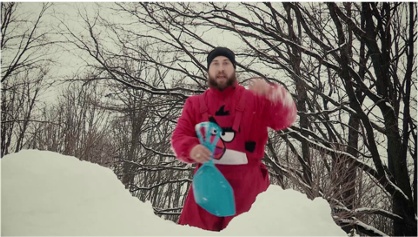}
    \captionsetup{font=footnotesize}
        \subcaption{Sampled Frame 3.}
        \label{fig: sf3}
    \end{minipage}%
    \begin{minipage}{0.25\textwidth}
        \centering
        \includegraphics[height=2cm]{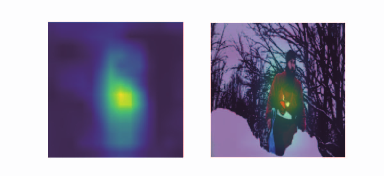}
        \includegraphics[height=2cm]{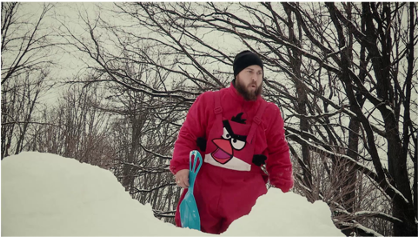}
    \captionsetup{font=footnotesize}
        \subcaption{Sampled Frame 4.}
        \label{fig: sf4}
    \end{minipage}

    \begin{minipage}{0.25\textwidth}
        \centering
        \includegraphics[height=2cm]{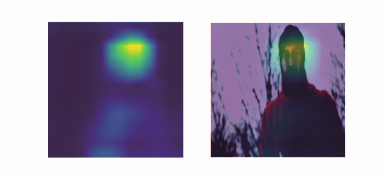}
        \includegraphics[height=2cm]{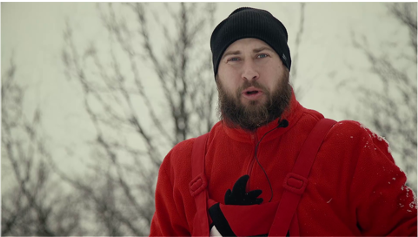}
    \captionsetup{font=footnotesize}
        \subcaption{Sampled Frame 5.}
        \label{fig: sf5}
    \end{minipage}%
    \begin{minipage}{0.25\textwidth}
        \centering
        \includegraphics[height=2cm]{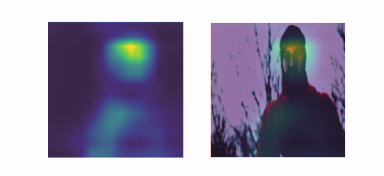}
        \includegraphics[height=2cm]{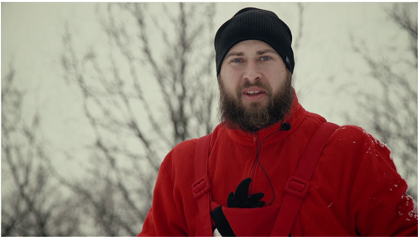}
    \captionsetup{font=footnotesize}
        \subcaption{Sampled Frame 6.}
        \label{fig: sf6}
    \end{minipage}%
    \begin{minipage}{0.25\textwidth}
        \centering
        \includegraphics[height=2cm]{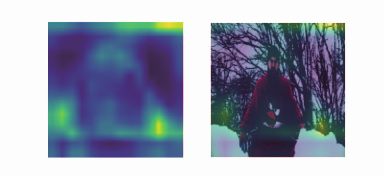}
        \includegraphics[height=2cm]{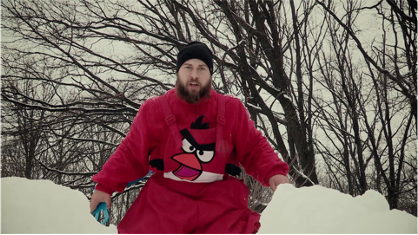}
    \captionsetup{font=footnotesize}
        \subcaption{Sampled Frame 7.}
        \label{fig: sf7}
    \end{minipage}%
    \begin{minipage}{0.25\textwidth}
        \centering
        \includegraphics[height=2cm]{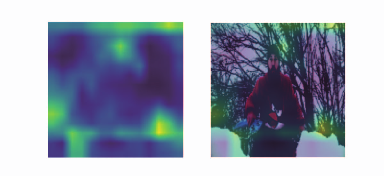}
        \includegraphics[height=2cm]{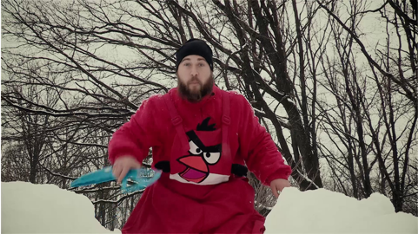}
    \captionsetup{font=footnotesize}
        \subcaption{Sampled Frame 8.}
        \label{fig: sf8}
    \end{minipage}

    \begin{minipage}{0.25\textwidth}
        \centering
        \includegraphics[height=2cm]{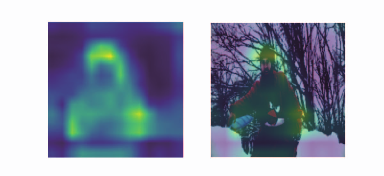}
        \includegraphics[height=2cm]{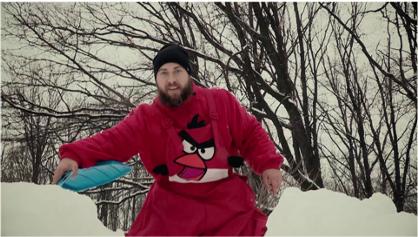}
    \captionsetup{font=footnotesize}
        \subcaption{Sampled Frame 9.}
        \label{fig: sf9}
    \end{minipage}%
    \begin{minipage}{0.25\textwidth}
        \centering
        \includegraphics[height=2cm]{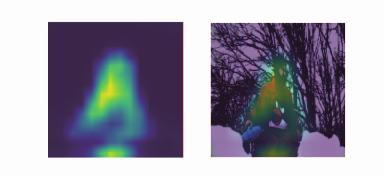}
        \includegraphics[height=2cm]{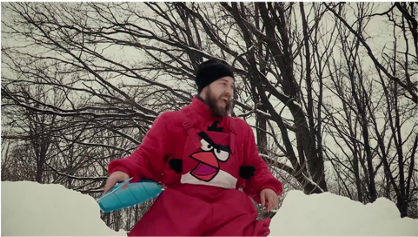}
    \captionsetup{font=footnotesize}
        \subcaption{Sampled Frame 10.}
        \label{fig: sf10}
    \end{minipage}%
    \begin{minipage}{0.25\textwidth}
        \centering
        \includegraphics[height=2cm]{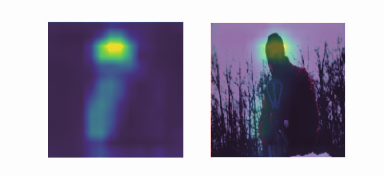}
        \includegraphics[height=2cm]{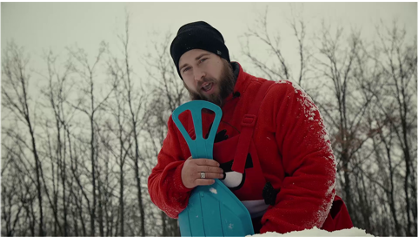}
    \captionsetup{font=footnotesize}
        \subcaption{Sampled Frame 11.}
        \label{fig: sf11}
    \end{minipage}%
    \begin{minipage}{0.25\textwidth}
        \centering
        \includegraphics[height=2cm]{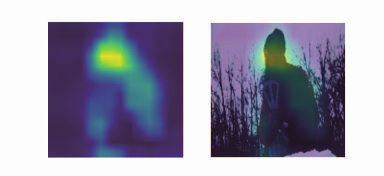}
        \includegraphics[height=2cm]{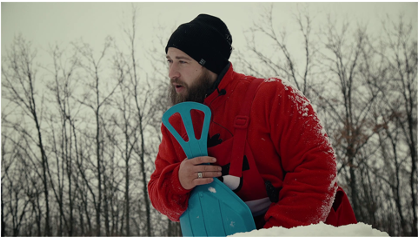}
    \captionsetup{font=footnotesize}
        \subcaption{Sampled Frame 12.}
        \label{fig: sf12}
    \end{minipage}

    \begin{minipage}{0.25\textwidth}
        \centering
        \includegraphics[height=2cm]{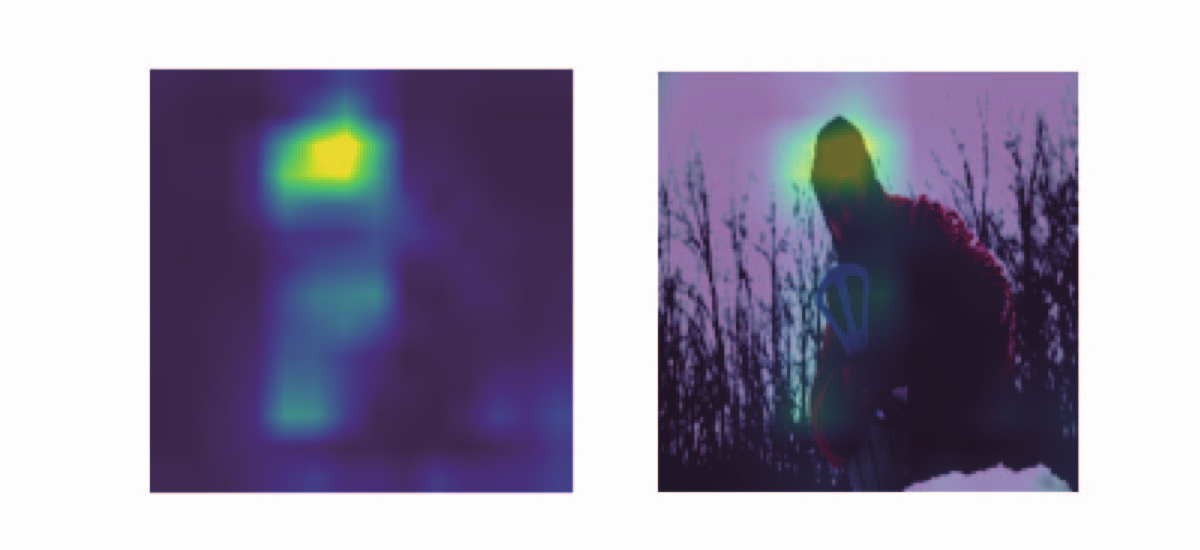}
        \includegraphics[height=2cm]{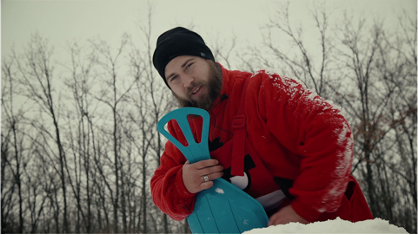}
    \captionsetup{font=footnotesize}
        \subcaption{Sampled Frame 13.}

    \end{minipage}%
    \begin{minipage}{0.25\textwidth}
        \centering
        \includegraphics[height=2cm]{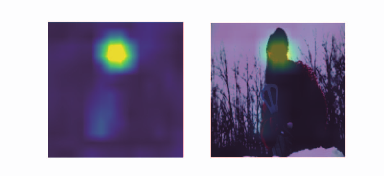}
        \includegraphics[height=2cm]{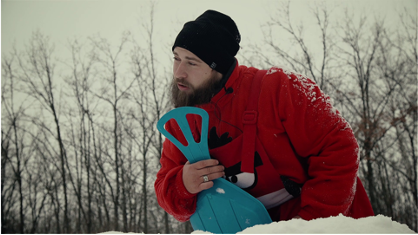}
    \captionsetup{font=footnotesize}
        \subcaption{Sampled Frame 14.}
        \label{fig: sf14}
    \end{minipage}%
    \begin{minipage}{0.25\textwidth}
        \centering
        \includegraphics[height=2cm]{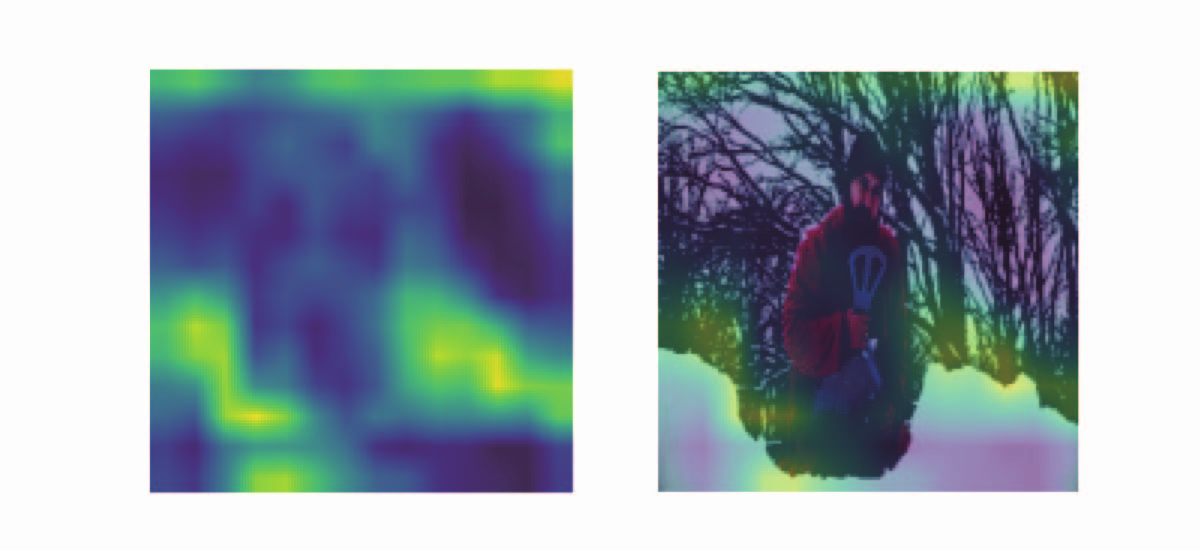}
        \includegraphics[height=2cm]{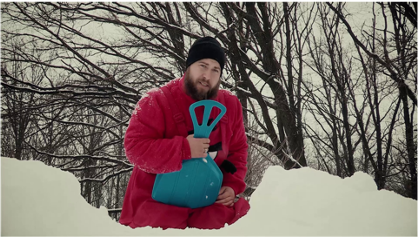}
    \captionsetup{font=footnotesize}
        \subcaption{Sampled Frame 15.}

    \end{minipage}%
    \begin{minipage}{0.25\textwidth}
        \centering
        \includegraphics[height=2cm]{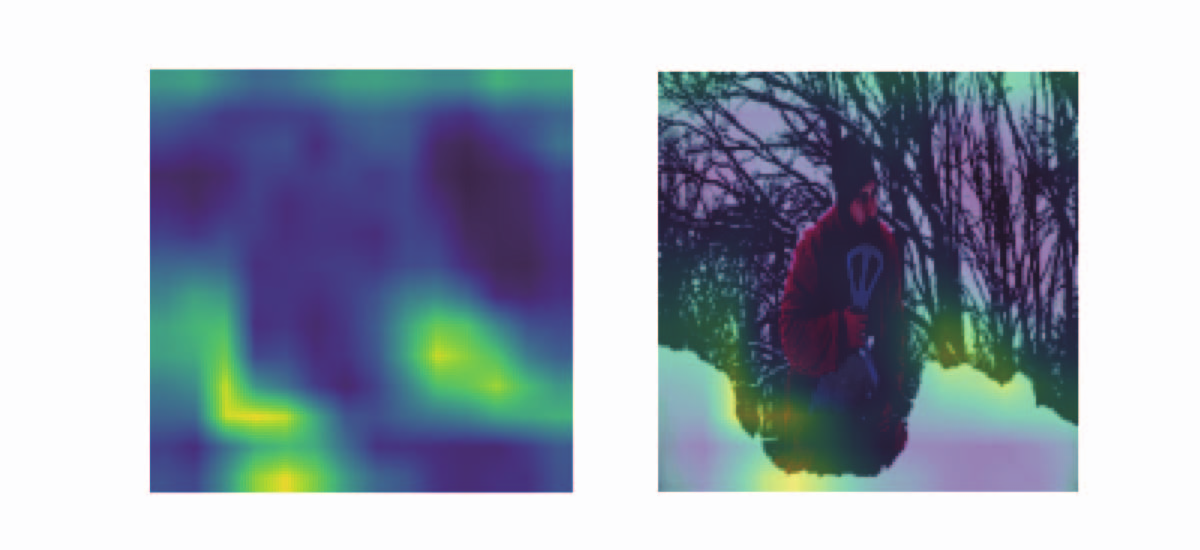}
        \includegraphics[height=2cm]{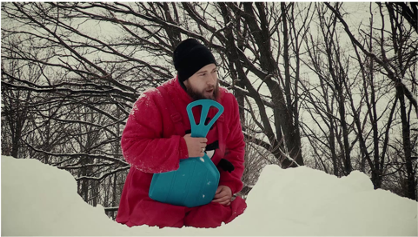}
    \captionsetup{font=footnotesize}
        \subcaption{Sampled Frame 16.}
        \label{fig: sf16}
    \end{minipage}

    \captionsetup{font=footnotesize}
    \caption{Original video frames and saliency visualisations (e.g. TelevisionClip\_1080P-68c6.mkv,  \textit{fps = 30}).}
    \label{fig: grid}
\end{figure}
\subsection{Introducing Saliency Features}
\label{subsec: introduce}
Based on the above, we designed experiments by fusing various features, namely NSS features and/or deep features with saliency. By incorporating saliency maps into video quality assessment models, our objective is to explore whether their perceptual relevance can be enhanced.

To address this limitation, we employed the Class Activation Maps (CAM)\cite{zhou2016learning}. In general, ConvNet consists of a series of convolutional layers, each of which is constructed as a set of filters, which are then connected to fully connected layers. The activation map represents the salient regions of an image for a specific prediction. Here we use Score-CAM~\cite{wang2020score} to implement feature activation maps, as this was one of the best performing methods.

Score-CAM generates a saliency map of an image based on the activation maps of the convolutional layers. The process involves extracting activation maps from an input image, using them as masks to linearly combine with corresponding output activation maps, and obtaining saliency maps through the ReLU function. The model implementation provides a pre-trained test model based on AlexNet~\cite{krizhevsky2012imagenet}, VGG-16~\cite{simonyan2014very}, and ResNet-18\cite{he2016deep}. In this paper, we apply the pre-trained model~\cite{wang2020score, torchvision2016} to obtain the saliency features of video frames. The implementation utilizes a pre-trained VGG-16 network as a baseline within the PyTorch framework. Input images are resized to ($224\times224\times3$) dimensions and normalized within the $[0, 1]$ range. These saliency maps were applied to sampled video frames (sampled at the frame rate of the video, fps), providing a finer understanding of the key regions influencing the quality prediction. 

\subsection{Fusion of Features}
\label{subsec: fusion}
We conducted a comprehensive study and deconstructed the baseline BVQA metrics. We explored different feature fusions and conducted ablation studies to explore their individual and combined effectiveness. The methodology consisted of the following steps. Firstly, we constructed different versions of BVQA metrics, each focusing on a specific set of features: traditional natural scene statistics features (NSS), deep features (DFs) extracted from pre-trained CNNs and content-aware feature extractors, as well as saliency maps. Then, we analysed the impact of these different features on video quality prediction by adding saliency features as new perceptual factors. Various feature combinations, such as NSS\_DFs, NSS\_saliency, and NSS\_DFs\_saliency, were considered, applicable to the entire dataset and spatial resolution groups. Additionally, we integrated saliency features into this fusion.

When integrating saliency features, we resized the frames of the test video sequences to match the model's input size ($224 \times 224$) and then performed feature extraction and fusion. Computing features for all frames in a video requires significant computational memory and time. To address this, we adopted a frame rate sampling approach for UGC videos, obtaining saliency maps for the sampled frames (one every sec). Here we have selected \textit{\textbf{TelevisionClip\_1080P-68c6.mkv}} from the YouTube-UGC dataset as an example, as shown in \textbf{Figure} \ref{fig: grid}. The video was then sampled at a frame rate (\textit{fps = 30}) to extract sample frames as example images for visualisation and different feature extraction tests. VSFACNN performs deep feature extraction on each frame of the video without frame sampling. Each frame of the video is fed into the Content-Aware Feature Extractor of the VSFA model to obtain deep semantic features resulting in a feature dimension of 4096. VSFACNN\_SALIENCY samples the video at a rate of one every second and extracts saliency feature maps and deep feature maps from the video frame. The two feature maps are then fused together by summing the corresponding positions in the two maps to create a new feature (dimension kept at 4096, consistent with VSFACNN). Both metrics require downsampling of the video to fit the input size of the pre-trained model.

\section{Experiments and Results}
\label{sec: res}
We first describe the experimental settings, including the datasets, the basic evaluation method and the environment configurations. Next, we present performance comparisons and analysis of prediction accuracy results for different combinations of feature metrics. Afterwards, we also demonstrate the trade-off between accuracy and complexity as a measure of computational efficiency. Then, we investigate the effect of saliency in detail with a correlation analysis of the association between the deep features and the saliency features.

\subsection{Experiment Setting}
\label{subsec: setting}
\textbf{UGC Datasets}. We use two state-of-the-art BVQA datasets for performance evaluation: the Konstanz Natural Video Database (KoNViD-1k)\cite{hosu2017konstanz} and YouTube User Generated Content Database(YouTube-UGC). We train and test separately but we also concatenate the two datasets (denoted as the COMBINED dataset) as an additional comprehensive benchmark. KoNViD-1k contains 1200 videos with a wide variety of video content. The time duration is 8 seconds at 24/25/30fps, the spatial resolution is 540p, and the MOS scale used is $[1, 5]$. YouTube-UGC is a collection of videos uploaded to the YouTube platform and is primarily used for video compression and quality assessment studies. It contains about 1500 videos across 14 genres (e.g., Animation, TelevisionClip). The time duration of each video is 20 seconds, the spatial resolutions include 360p, 480p, 720p, 1080p, and 2160p, and the subjective scores as within the same MOS range of KoNViD-1k. Due to the computational resource limitations of our experimental environment, we did not use 4K videos in this work, which resulted in a total of 1039 videos in 360p (224), 480p (265), 720p (256), and 1080p (294) from all categories (except VR). We also find that Sports, Vlogs and Gaming are the three categories with the largest share in the YouTube-UGC dataset. This fully reflects the present focus of UGC, i.e., live games, sports, and self-publishing being the mainstream videos.

\textbf{Evaluation Method}. 
For performance evaluation, we use SVR to predict subjective scores (MOS). For the training of the SVRs, 100 iterations of hold-out cross-validation random splits (train:test = 80\%:20\% split) were executed. Optimal SVR hyperparameters were selected using grid search tuning on a random 20\% of the training set. For the assessment of the BVQA metrics performance with and without saliency, we employed evaluation metrics such as Spearman's rank correlation coefficient (SRCC), Kendall's rank correlation coefficient (KRCC), Pearson's linear correlation coefficient (PLCC), and Root Mean Square Error (RMSE) are used to assess the monotonicity, linearity, and accuracy of the predictions. In this case, PLCC and RMSE are calculated after a nonlinear four-parameter logistic regression\cite{seshadrinathan2010study} as below, where $\beta$ parameters are fitted using least squares:
\begin{equation}
\label{eq: fx}
f(x) = \beta_2 + \frac{\beta_1 - \beta_2}{1 + e^{-x + \frac{\beta_3}{|\beta_4|}}} \, ,
\end{equation}
\begin{table}[ht]
    \caption{Performance comparison of the two UGC video datasets and the combined one. We applied 100 iterations of 80-20 hold-out cross-validation. The \underline{{\bfseries asterisked*}}, \underline{{\bfseries underlined}}, and {\bfseries boldfaced} entries indicate the best, top-2, and top-3, respectively.}
    \label{tab:methodcompare}
    \centering
    \begin{subtable}[t]{0.85\textwidth}
        \centering
        \begin{adjustbox}{width=\textwidth}
        \begin{tabular}{*{6}{c}}
            \toprule[1pt]
            \toprule[1pt]
            \multirowcell{2}{DATASET\\METRICS} & \multicolumn{4}{c}{KoNViD-1k} \\
            \cmidrule(lr){2-5}
            & $SRCC(\pm std)\uparrow$ & $KRCC(\pm std)\uparrow$ & $PLCC(\pm std)\uparrow$ & $RMSE(\pm std)\downarrow$ \\
            \midrule 
            SALIENCY & 0.1729(±0.0727) & 0.1167(±0.0487) & 0.1983(±0.0543) & 0.6263(±0.0195) \\
            NSS & 0.7747(±0.0242) & 0.5776(±0.0233) & 0.7723(±0.0247) & 0.4034(±0.0204) \\
            NSS\_SALIENCY & 0.7676(±0.0241) & 0.5716(±0.0221) & 0.7700(±0.0229) & 0.4082(±0.0186) \\
            \hdashline[0.8pt/2pt]
            NSS\_CNN (RAPIQUE) & \underline{{\bfseries0.8042(±0.0230)}} & 0.6133(±0.0226) & 0.8150(±0.0203) & 0.3702(±0.0169) \\
            NSS\_CNN\_SALIENCY & {\bfseries0.8013(±0.0228)} & 0.6088(±0.0220) & 0.8122(±0.0204) & 0.3735(±0.0172) \\
            NSS\_VSFACNN & 0.7945(±0.0249) & 0.6034(±0.0242) & 0.8023(±0.0229) & 0.3790(±0.0175) \\
            NSS\_VSFACNN\_SALIENCY & 0.7944(±0.0248) & 0.6035(±0.0241) & 0.8022(±0.0229) & 0.3792(±0.0175) \\
            \hdashline[0.8pt/2pt]
            CNN (ResNet-50) & 0.7116(±0.0320) & 0.5253(±0.0281) & 0.7477(±0.0276) & 0.4233(±0.0170) \\
            CNN\_SALIENCY & 0.7130(±0.0328) & 0.5255(±0.0290) & 0.7468(±0.0277) & 0.4254(±0.0173) \\
            VSFACNN & \underline{{\bfseries0.8059(±0.0234)*}} & 0.6141(±0.0237) & 0.8155(±0.0209) & 0.3692(±0.0171) \\
            VSFACNN\_SALIENCY & 0.7225(±0.0338) & 0.5317(±0.0296) & 0.7476(±0.0290) & 0.4235(±0.0178) \\
            \hdashline[0.8pt/2pt]
            UVQ\cite{wang2021rich} & 0.7089 & 0.5124 & 0.7054 & 0.4835 \\
            \bottomrule[1pt]
        \end{tabular}
        \end{adjustbox}
    \end{subtable}

    \begin{subtable}[t]{0.85\textwidth}
        \centering
        \begin{adjustbox}{width=\textwidth}
        \begin{tabular}{*{6}{c}}
            \toprule[1pt]
            \multirowcell{2}{DATASET\\METRICS} & \multicolumn{4}{c}{YouTube-UGC} \\
            \cmidrule(lr){2-5}
            & $SRCC(\pm std)\uparrow$ & $KRCC(\pm std)\uparrow$ & $PLCC(\pm std)\uparrow$ & $RMSE(\pm std)\downarrow$ \\
            \midrule 
            SALIENCY & 0.2048(±0.0707) & 0.1386(±0.0487) & 0.2574(±0.0674) & 0.6163(±0.0293) \\
            NSS & 0.7103(±0.0361) & 0.5249(±0.0305) & 0.7042(±0.0378) & 0.4526(±0.0267) \\
            NSS\_SALIENCY & 0.7038(±0.0394) & 0.5177(±0.0335) & 0.7044(±0.0392) & 0.4493(±0.0252) \\
            \hdashline[0.8pt/2pt]
            NSS\_CNN (RAPIQUE) & \underline{{\bfseries0.7546(±0.0332)}} & 0.5624(±0.0306) & 0.7667(±0.0318) & 0.4143(±0.0214) \\
            NSS\_CNN\_SALIENCY & {\bfseries0.7541(±0.0342)} & 0.5603(±0.0308) & 0.7637(±0.0325) & 0.4145(±0.0214) \\
            NSS\_VSFACNN & 0.7338(±0.0352) & 0.5445(±0.0317) & 0.7510(±0.0324) & 0.4194(±0.0210) \\
            NSS\_VSFACNN\_SALIENCY & 0.7333(±0.0350) & 0.5431(±0.0315) & 0.7499(±0.0327) & 0.4190(±0.0210) \\
            \hdashline[0.8pt/2pt]
            CNN (ResNet-50) & 0.6438(±0.0410) & 0.4655(±0.0336) & 0.6784(±0.0377) & 0.4718(±0.0211) \\
            CNN\_SALIENCY & 0.6274(±0.0452) & 0.4519(±0.0364) & 0.6651(±0.0412) & 0.4782(±0.0230) \\
            VSFACNN & \underline{{\bfseries0.7918(±0.0275)*}} & 0.5951(±0.0271) & 0.7977(±0.0259) & 0.3815(±0.0208) \\
            VSFACNN\_SALIENCY & 0.6441(±0.0428) & 0.4640(±0.0355) & 0.6805(±0.0375) & 0.4702(±0.0189) \\
            \hdashline[0.8pt/2pt]
            UVQ\cite{wang2021rich} & - & - & - & - \\
            \bottomrule[1pt]
        \end{tabular}
        \end{adjustbox}
    \end{subtable}

    \begin{subtable}[t]{0.85\textwidth}
        \centering
        \begin{adjustbox}{width=\textwidth}
        \begin{tabular}{*{6}{c}}
            \toprule[1pt]
            \multirowcell{2}{DATASET\\METRICS} & \multicolumn{4}{c}{COMBINED} \\
            \cmidrule(lr){2-5}
            & $SRCC(\pm std)\uparrow$ & $KRCC(\pm std)\uparrow$ & $PLCC(\pm std)\uparrow$ & $RMSE(\pm std)\downarrow$ \\
            \midrule 
            SALIENCY & 0.1982(±0.0440) & 0.1339(±0.0298) & 0.2273(±0.0463) & 0.6826(±0.0195) \\
            NSS & 0.7556(±0.0245) & 0.5625(±0.0219) & 0.7546(±0.0241) & 0.4603(±0.0184) \\
            NSS\_SALIENCY & 0.7492(±0.0199) & 0.5569(±0.0173) & 0.7545(±0.0183) & 0.4613(±0.0148) \\
            \hdashline[0.8pt/2pt]
            NSS\_CNN (RAPIQUE) & \underline{{\bfseries0.7839(±0.0200)}} & 0.5942(±0.0188) & 0.8020(±0.0175) & 0.4189(±0.0153) \\
            NSS\_CNN\_SALIENCY & {\bfseries0.7815(±0.0214)} & 0.5911(±0.0199) & 0.7992(±0.0185) & 0.4212(±0.0155) \\
            NSS\_VSFACNN & 0.7756(±0.0217) & 0.5839(±0.0203) & 0.7944(±0.0177) & 0.4239(±0.0147) \\
            NSS\_VSFACNN\_SALIENCY & 0.7755(±0.0213) & 0.5835(±0.0198) & 0.7942(±0.0175) & 0.4245(±0.0145) \\
            \hdashline[0.8pt/2pt]
            CNN (ResNet-50) & 0.6984(±0.0255) & 0.5082(±0.0222) & 0.7377(±0.0218) & 0.4741(±0.0150) \\
            CNN\_SALIENCY & 0.6878(±0.0269) & 0.4999(±0.0230) & 0.7330(±0.0222) & 0.4786(±0.0150) \\
            VSFACNN & \underline{{\bfseries0.7978(±0.0180)*}} & 0.6044(±0.0180) & 0.8162(±0.0152) & 0.4063(±0.0124) \\
            VSFACNN\_SALIENCY & 0.6986(±0.0265) & 0.5109(±0.0231) & 0.7326(±0.0213) & 0.4741(±0.0140) \\
            \hdashline[0.8pt/2pt]
            UVQ\cite{wang2021rich} & - & - & - & - \\
            \bottomrule[1pt]
        \end{tabular}
        \end{adjustbox}
    \end{subtable}
\end{table}

\textbf{Environment Setting}. 
All experiments were executed on the BlueCrystal4 platform at the Advanced Computing Research Centre, University of Bristol, equipped with 525 Lenovo nx360 m5 compute nodes, each with two 14-core 2.4 GHz Intel E5-2680 v4 (Broadwell) CPUs and 128 GiB of RAM. All models were implemented under the Ubuntu system with Matlab R2018a and Python 3.7/Python3.6. 
\begin{figure}[ht]
    \begin{subfigure}{0.3\linewidth}
        \centering
        \includegraphics[width=\linewidth]{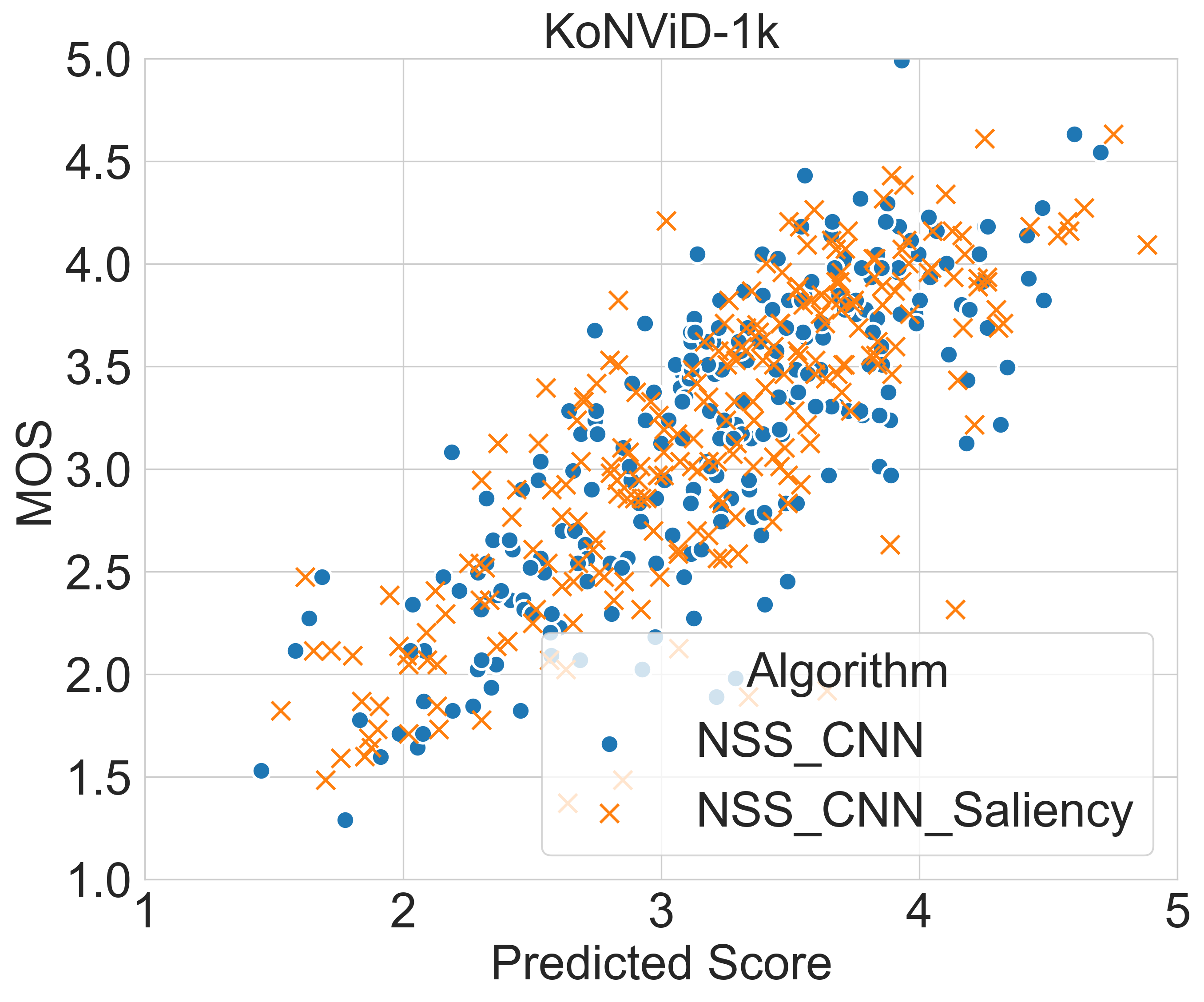}
    \end{subfigure}
    \begin{subfigure}{0.3\linewidth}
        \centering
        \includegraphics[width=\linewidth]{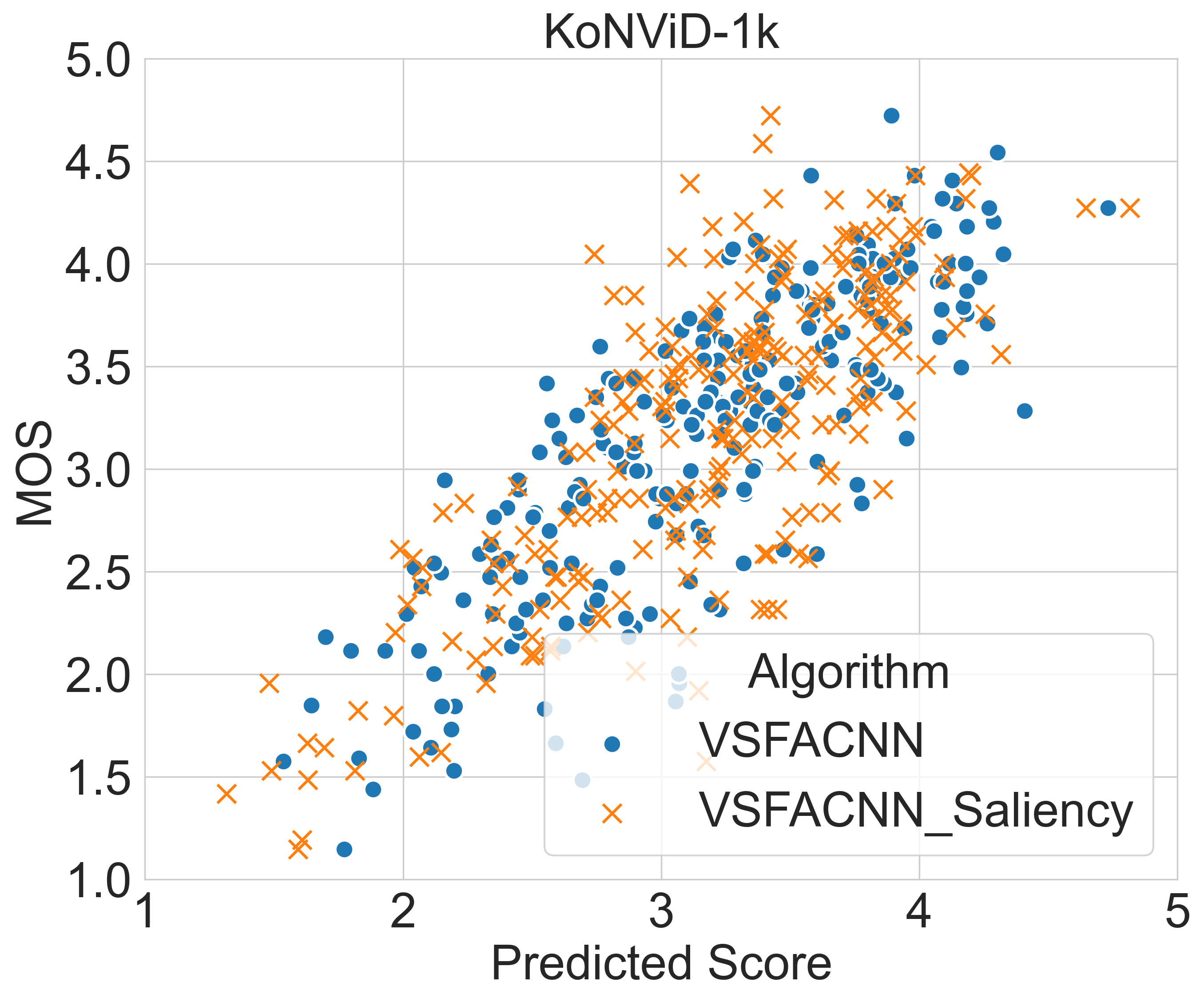}
    \end{subfigure}
    \begin{subfigure}{0.3\linewidth}
        \centering
        \includegraphics[width=\linewidth]{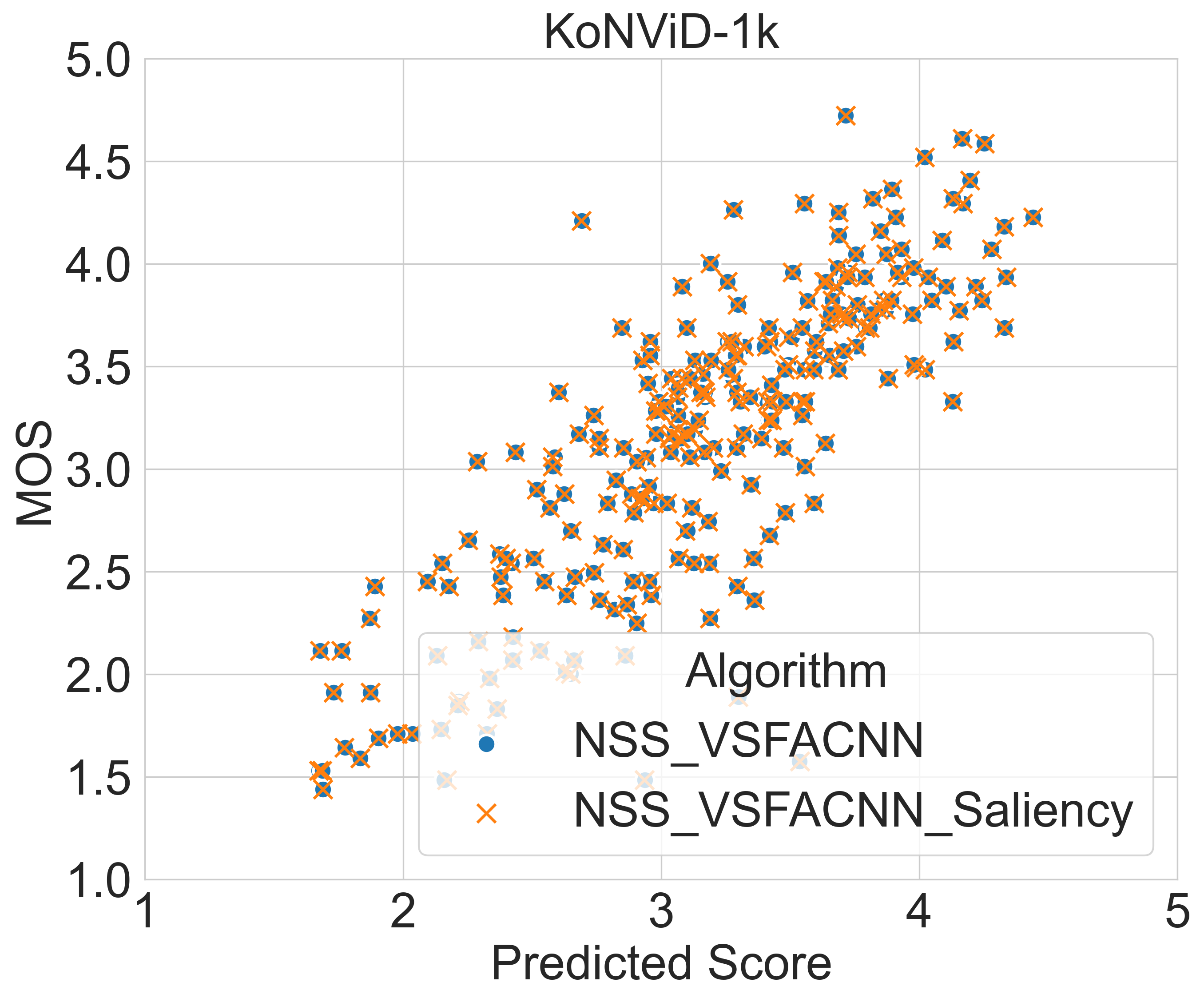}
    \end{subfigure}
    
    \centering
    \begin{subfigure}{0.3\linewidth}
        \centering
        \includegraphics[width=\linewidth]{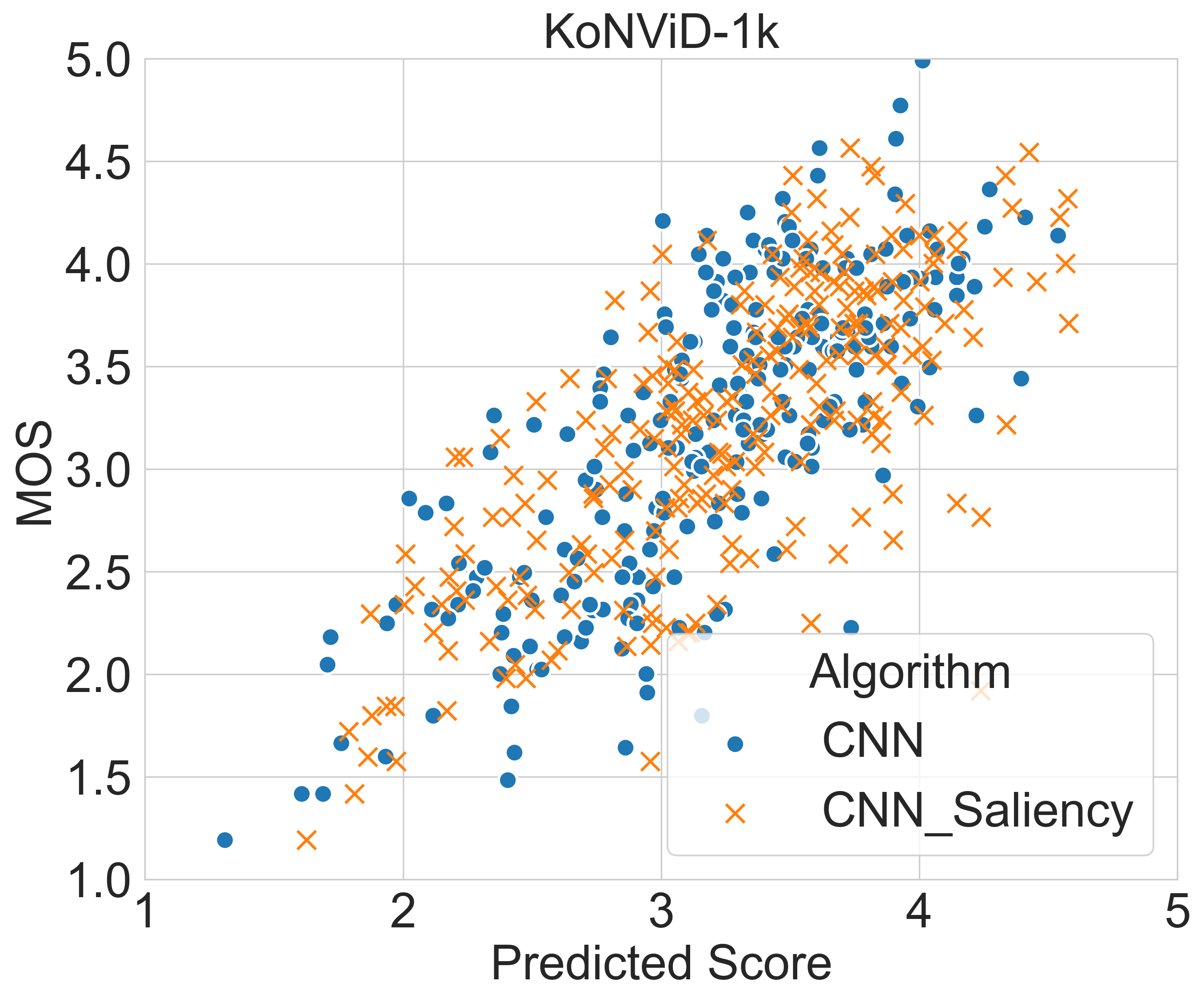}
    \end{subfigure}
    \begin{subfigure}{0.3\linewidth}
        \centering
        \includegraphics[width=\linewidth]{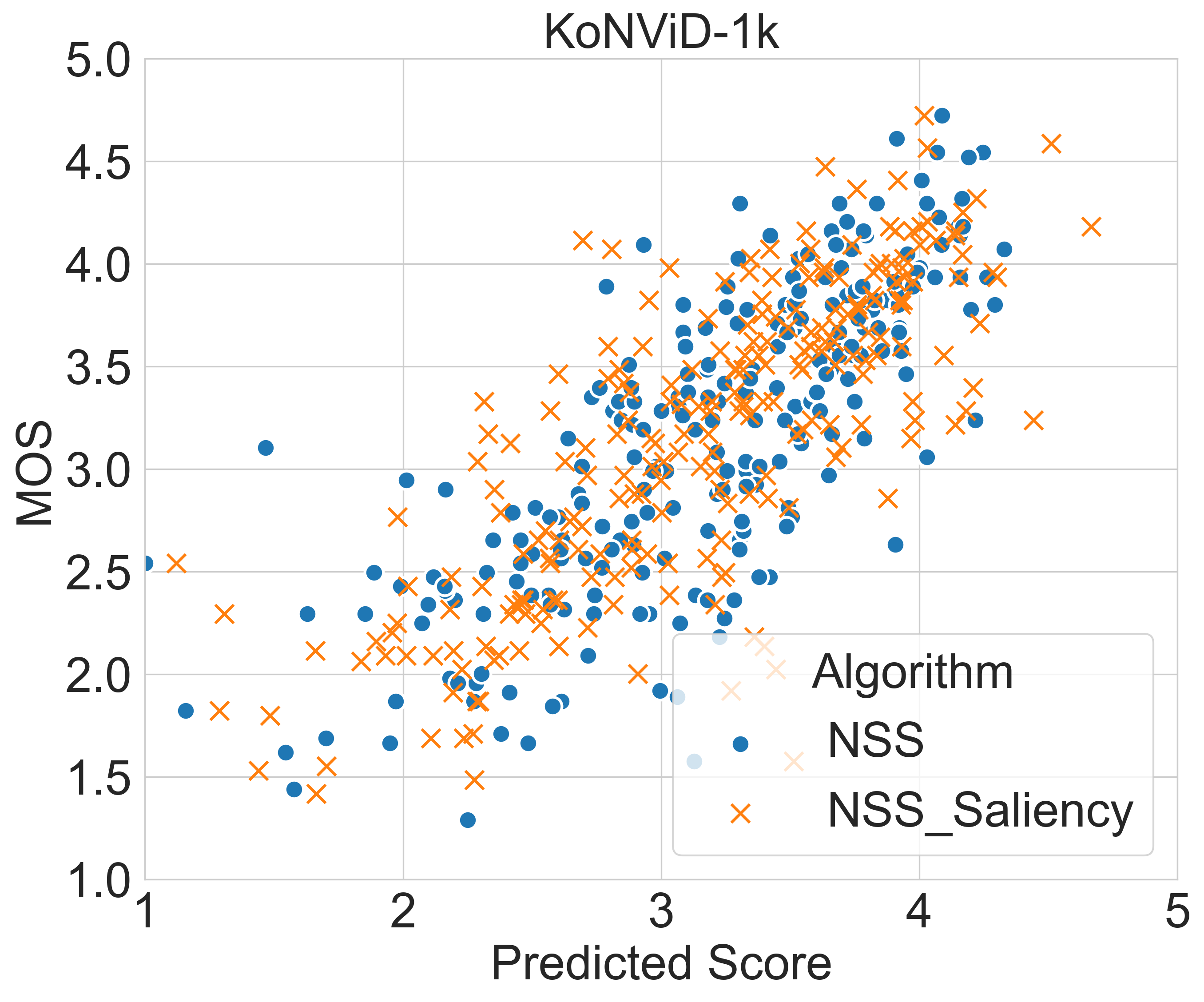}
    \end{subfigure}
    \captionsetup{font=footnotesize}
    \caption{Scatter plots of BVQA metrics versus MOS trained with a grid-search SVR using cross-validation on the KoNViD-1k Dataset. Similar observations can be made for the other datasets.}
    \label{fig:konvid}
\end{figure}
\begin{figure}[ht]
    \begin{subfigure}{0.3\linewidth}
        \centering
        \includegraphics[width=\linewidth]{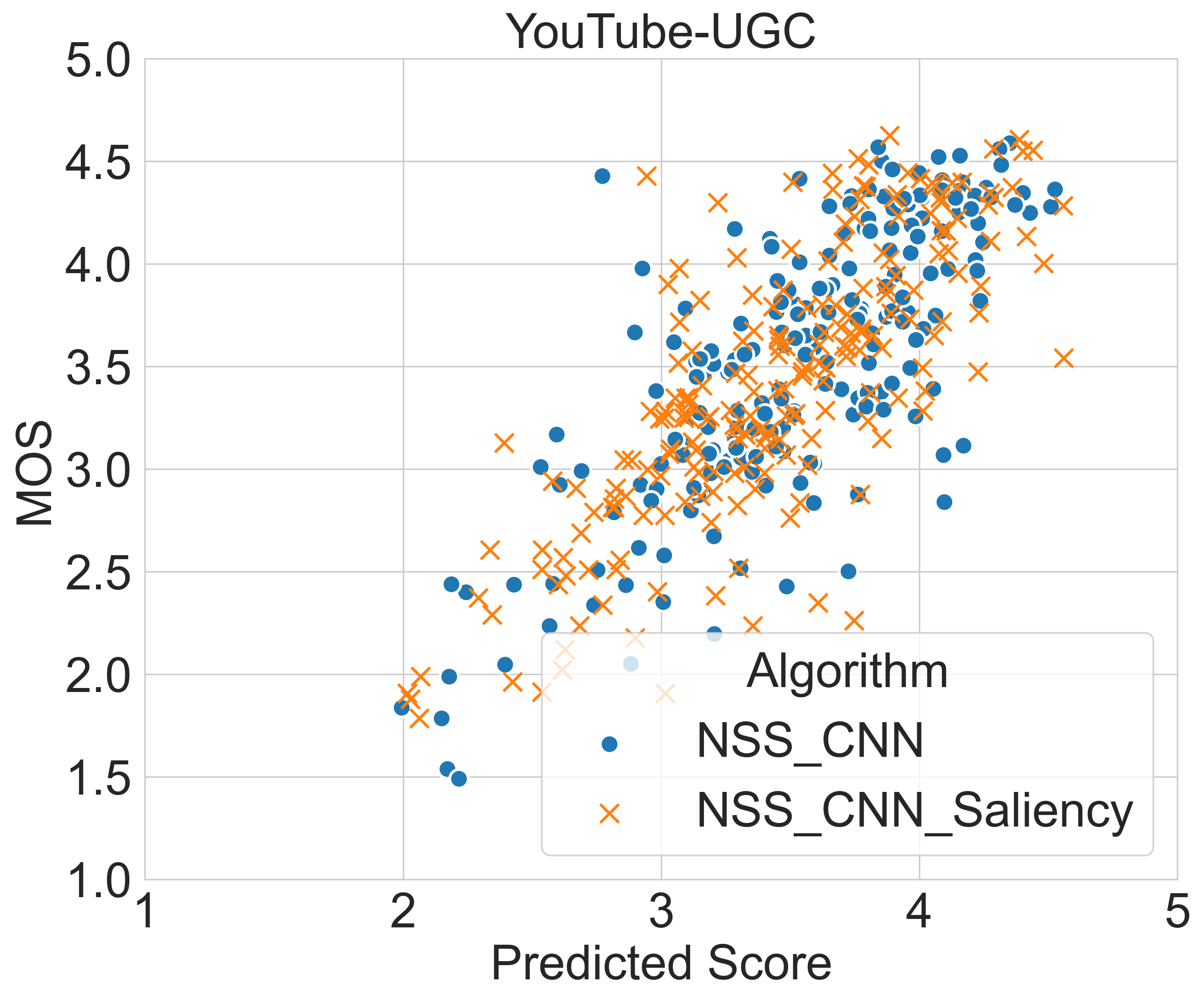}
    \end{subfigure}
    \begin{subfigure}{0.3\linewidth}
        \centering
        \includegraphics[width=\linewidth]{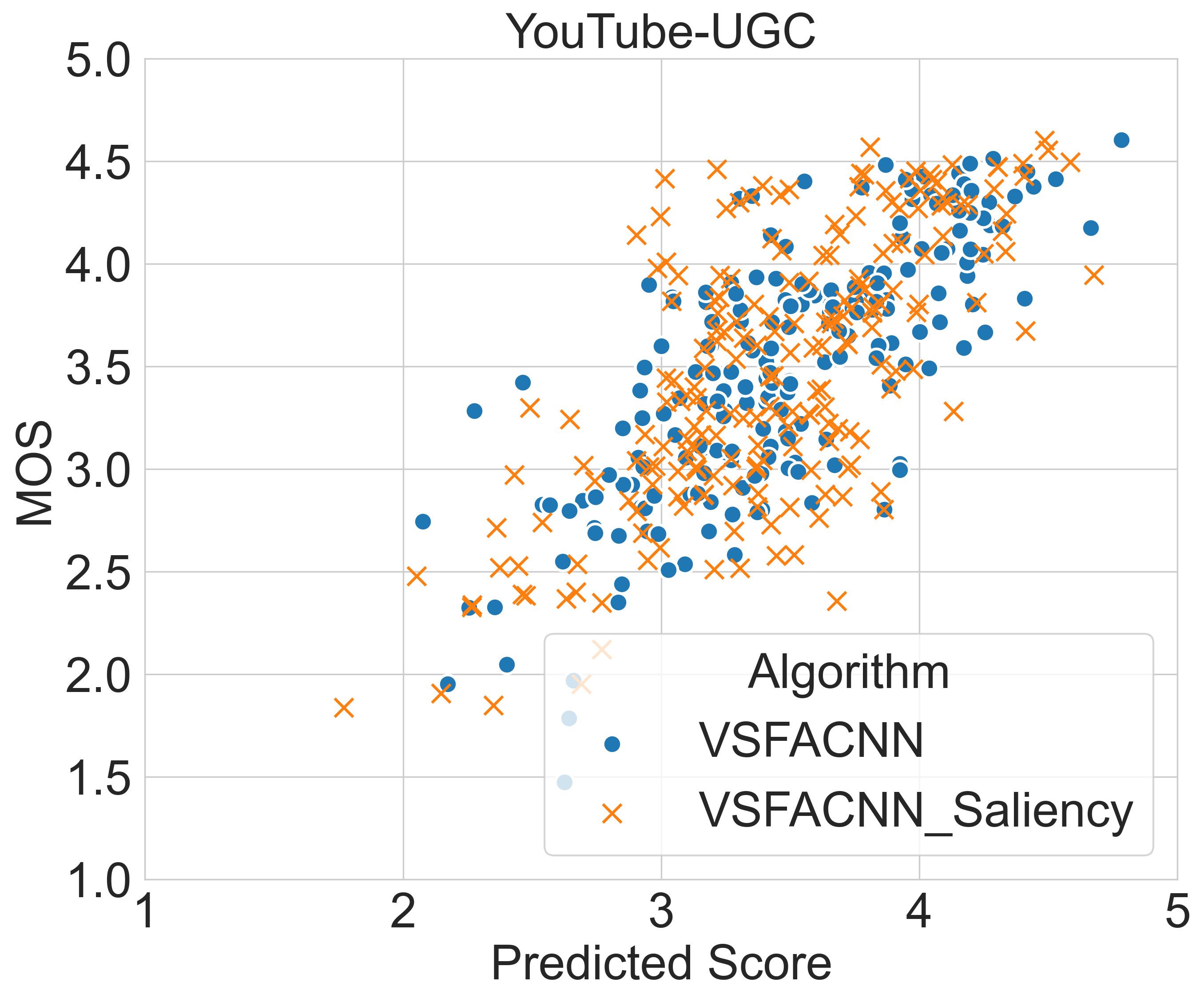}
    \end{subfigure}
    \begin{subfigure}{0.3\linewidth}
        \centering
        \includegraphics[width=\linewidth]{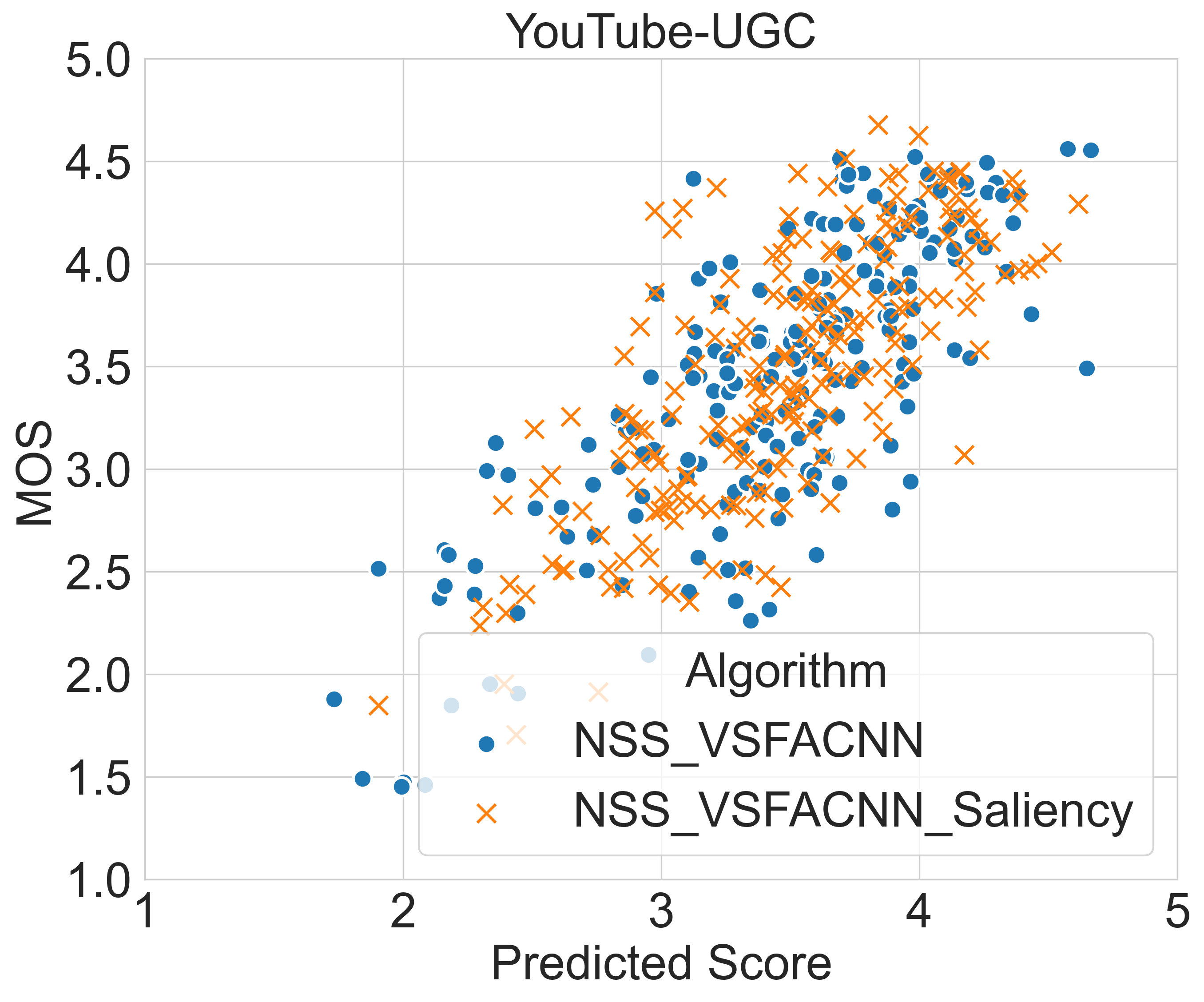}
    \end{subfigure}
    \centering
    \begin{subfigure}{0.3\linewidth}
        \centering
        \includegraphics[width=\linewidth]{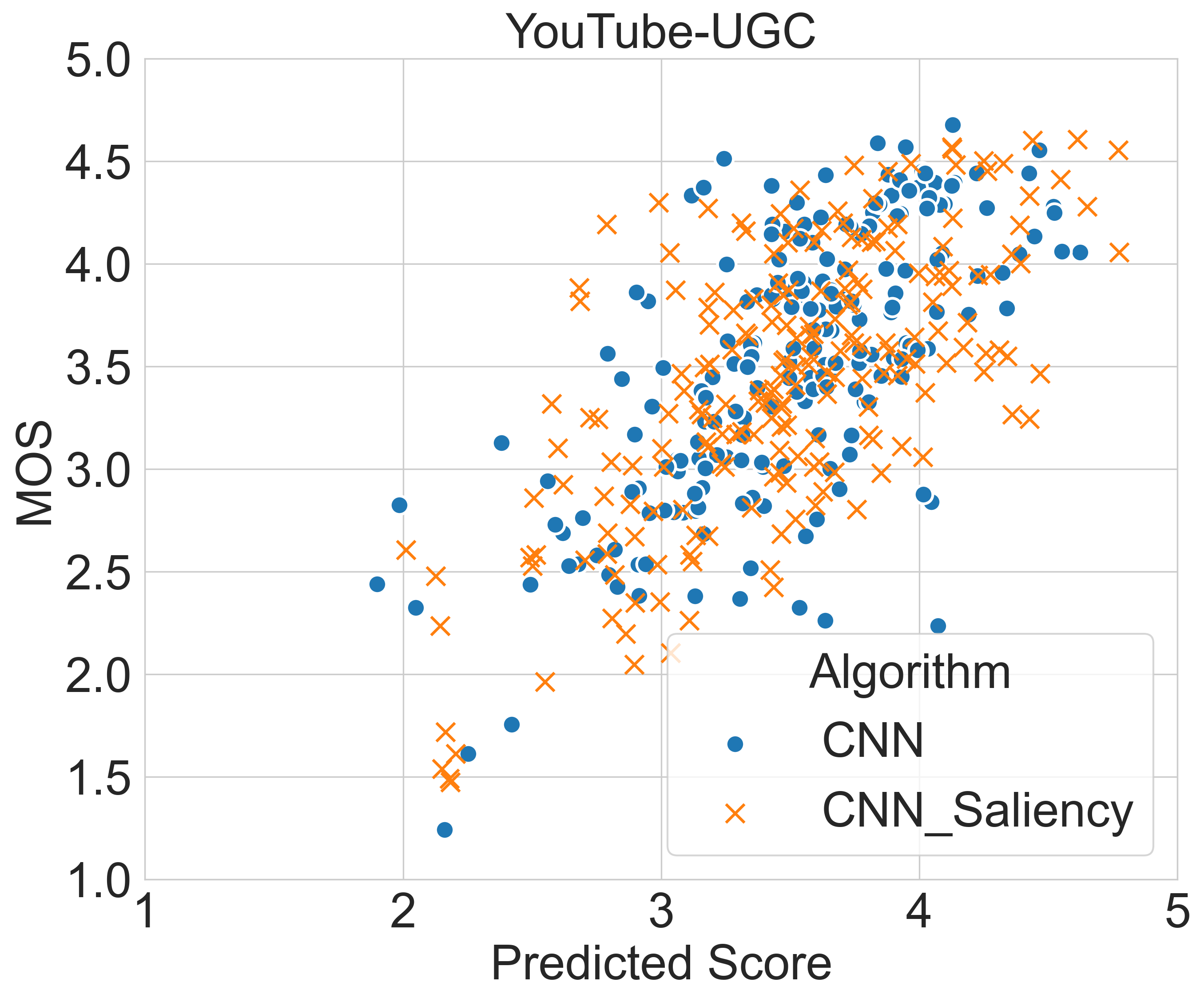}
    \end{subfigure}
    \begin{subfigure}{0.3\linewidth}
        \centering
        \includegraphics[width=\linewidth]{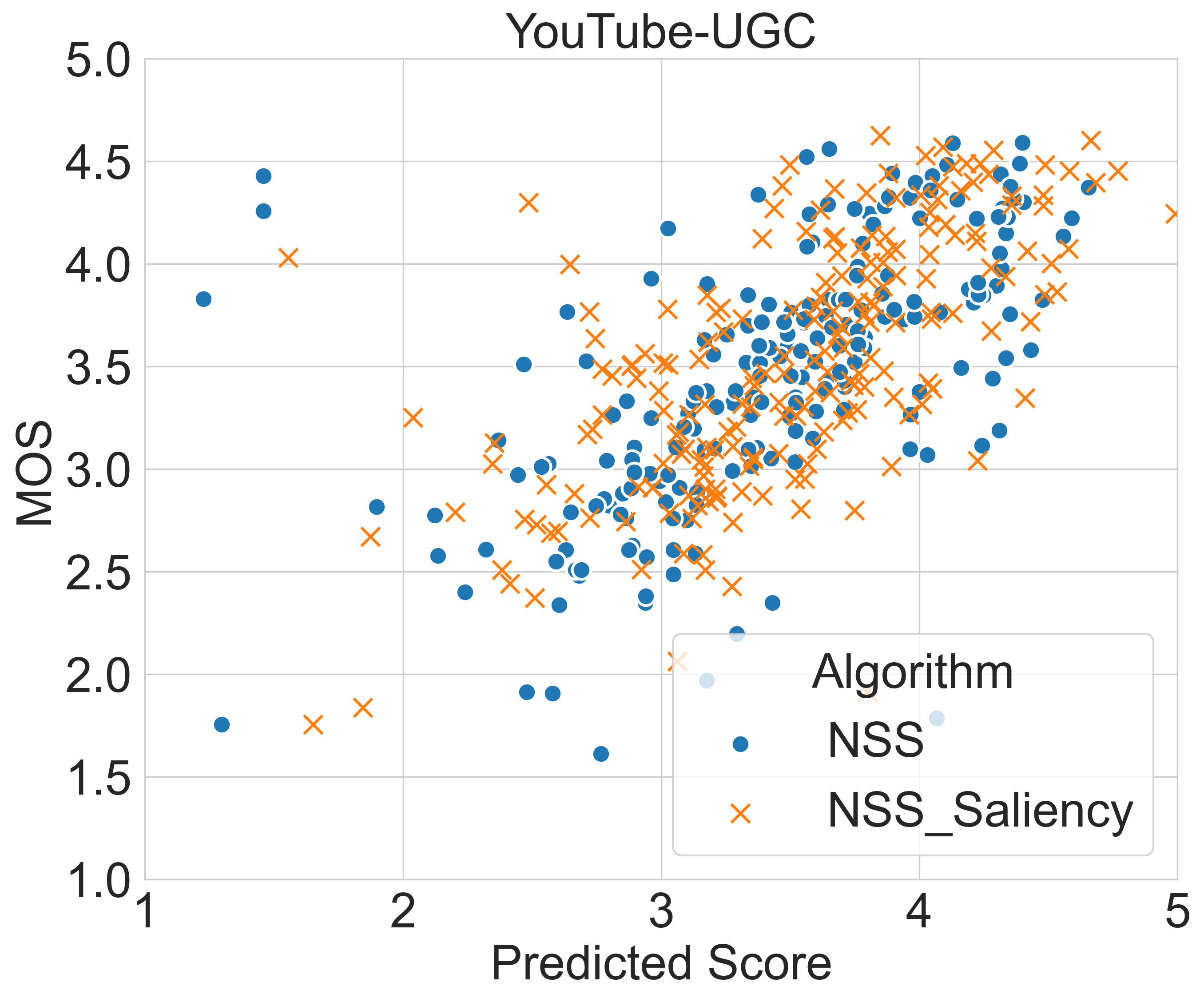}
    \end{subfigure}    
    \captionsetup{font=footnotesize}
    \caption{Scatter plots of BVQA metrics versus MOS trained with a grid-search SVR using cross-validation on the YouTube-UGC Dataset. Similar observations can be made for the other datasets.}
    \label{fig:youtube}
\end{figure}
\subsection{Performance Comparison}
\label{subsec: performance}
We collected the results of the tested metrics along with the feature combinations with and without saliency and summarised them in \textbf{Table}~\ref{tab:methodcompare}.
This table shows the performance comparison results for the three evaluation datasets in subtables based on the correlation metrics and the RMSE. We need to note that the figures reported here are the median values across all executions. 
In these subtables, we also included results for a recently released model, called  UVQ~\cite{Wang_2021_CVPR}, a rich feature for perceptual quality assessment, tested only on KoNViD-1k due to the currently publicly available pre-trained baseline model trained on YT-UGC$^+$. In addition to the table, we selectively illustrate results in groups of scatterplots in \textbf{Figures} \ref{fig:konvid} and  \textbf{Figures} \ref{fig:youtube}. 

From \textbf{Table}~\ref{tab:methodcompare} and \textbf{Figures} \ref{fig:konvid}-\ref{fig:youtube}, we observe that VSFACNN exhibits the best performance across all datasets, followed by NSS\_CNN (RAPIQUE). By using only extracted deep features to predict video quality scores with grid search SVR and cross-validation, it achieves high prediction accuracy without even modelling long-term dependencies. It can be seen that the best performance is achieved on KoNViD-1k using only VSFACNN. The results show that improved SRCC with subjective scores can be achieved by using deep spatial features alone extracted with a VSFA ($\Delta$ SRCC = 0.0372, $\Delta$ RMSE = 0.0328 over the state-of-the-art metric RAPIQUE~\cite{tu2021rapique}). Moreover, we observed that incorporating saliency features into both RAPIQUE and VSFA metrics resulted in predictions with comparable accuracy to the original models. It is therefore inferred that the addition of saliency does not further improve the performance. 

As can be seen in \textbf{Figure} \ref{fig:konvid} and \textbf{Figure} \ref{fig:youtube}, by comparing the metrics with and without the saliency feature, we observe that introducing saliency results in a bigger deviation for some points. It is noteworthy that in \textbf{Figure} \ref{fig:konvid}, both NSS\_VSFACNN and NSS\_VSFACNN\_SALIENCY demonstrate nearly indistinguishable performances on the KoNViD-1k dataset. These results suggest that the hypothesis we made that the saliency maps would improve the metrics' perceptual alignment is not valid. A possible explanation could be that some of the features examined and particularly the DF could already be implicitly representing saliency. In the next sections, we perform an investigation of the topic.

\subsection{Accuracy-Complexity Trade-offs}
\label{subsec: complexity}
In addition to performance, computational efficiency is critical for BVQA metrics. \textbf{Table} \ref{tab:tradeoff} shows a comparison of the metrics' dimensions and runtime, where we analyse the trade-off between prediction accuracy and computational efficiency. The feature extraction time refers to extracting a feature vector of one video sequence, while the quality prediction time refers to the running time of SVR. We selected ten 1080p videos from the YouTube-UGC dataset, covering all video categories. All runtime tests were conducted on the BlueCrystal4 server at the University of Bristol, with the feature extraction runtimes represented as the average time in seconds. 

\begin{figure}[ht]
    \begin{minipage}{0.5\linewidth}
        \centering
        \includegraphics[height=6cm]{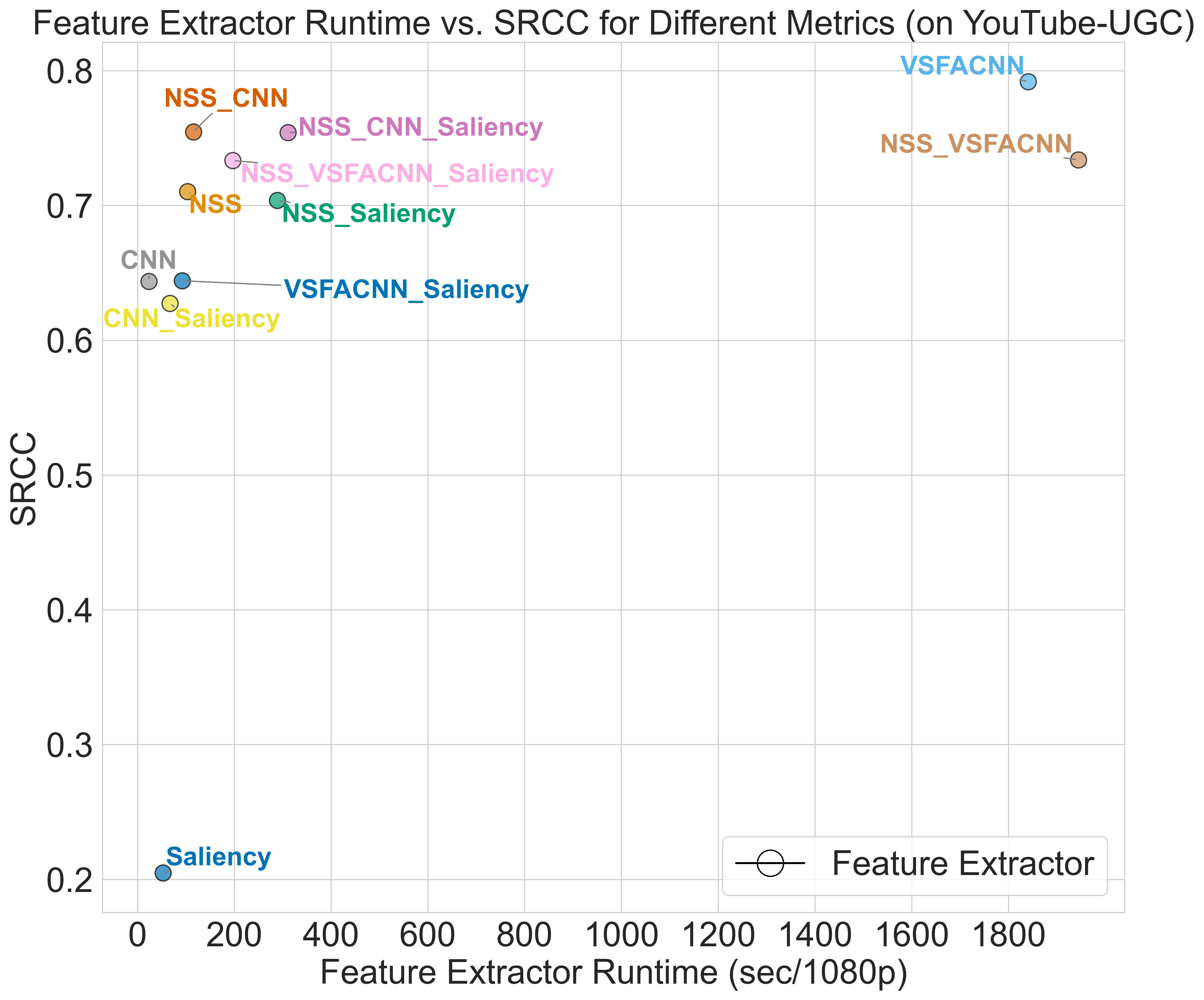}
    \end{minipage}%
    \begin{minipage}{0.5\linewidth}
        \centering
        \includegraphics[height=6cm]{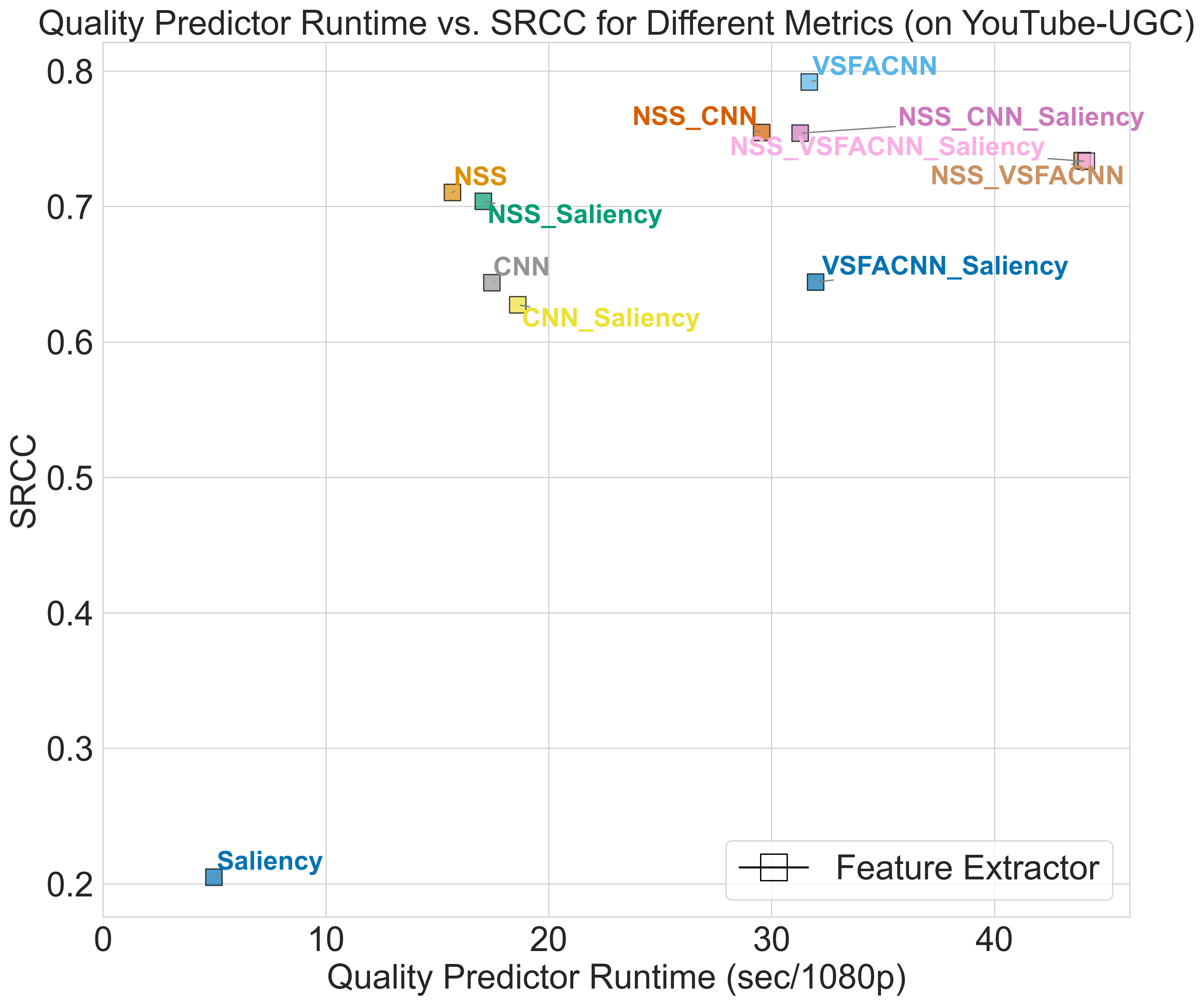}
    \end{minipage}
    \captionsetup{font=footnotesize}
    \caption{Accuracy-Complexity Comparison on YouTube-UGC (across ten 1080p test videos).}
    \label{fig:runtime}
\end{figure}

As anticipated, incorporating saliency into the feature extraction process increases the runtime, but it does not improve the prediction accuracy. Particularly, the runtime for SALIENCY feature extraction is almost half of the runtime for NSS feature extraction. Furthermore, we observed from \textbf{Figure} \ref{fig:runtime} that the combination of VSFACNN with metrics results in the longest runtime, approximately 16 times longer than RAPIQUE. The reason is that VSFACNN performs deep feature extraction on each frame of the video without frame sampling, resulting in high accuracy at the expense of significant runtime. In contrast, VSFACNN\_SALIENCY samples the video at a rate of one frame at every second and downscales them to lower resolutions ($224 \times 224$) for extracting both video deep and saliency features at the same time. Therefore, it demonstrates a shorter runtime and simultaneously relatively lower accuracy. 
\begin{table}[ht]
    \caption{Comparison of accuracy, dimensionality and runtime on different metrics (seconds, approx.). The evaluation was performed on ten 1080p test videos (from different genres) in the YouTube-UGC dataset. Runtime results are averaged over the 1080p test dataset.}
    \label{tab:tradeoff}
    \centering
    \begin{tabular}{*{14}{r}}
    \toprule[1pt]
    \toprule[1pt]
        \multirowcell{2}{METRICS} & \multirowcell{2}{DIM} & \multicolumn{2}{c}{RUNTIME}\\ 
        \cmidrule(lr){3-4} 
        && Feature Extractor & Quality Predictor \\
        \midrule 

        SALIENCY 
        &224	&52.62	&4.98\\
        NSS 
        &1836	&103.31	&15.68\\
        NSS\_SALIENCY 
        &2060	&289.22	&17.06\\
        NSS\_CNN (RAPIQUE) 
        &3884	&115.36	&29.56\\
        NSS\_CNN\_SALIENCY 
        &4108	&310.81	&31.28\\
        NSS\_VSFACNN 
        &5932	&1944.48	&43.94\\
        NSS\_VSFACNN\_SALIENCY 
        &5932	&196.73	&44.13\\
        CNN (ResNet-50) 
        &2048	&23.74	&17.45\\
        CNN\_SALIENCY 
        &2272	&66.99	&18.61\\
        VSFACNN 
        &4096	&1841.13	&31.69\\
        VSFACNN\_SALIENCY 
        &4096	&92.27	&31.98\\

        \bottomrule[1pt]
    \end{tabular}
\end{table}
Additionally, we found that its accuracy is comparable to ResNet-50 (CNN), which further indicates that saliency is likely to be represented by the features extracted by deep neural networks.

\subsection{Deep Features Visualisation}
\label{subsec: visulisa}
We visualise deep network features and saliency features to gain insight into the inherent salient regions in video frames embodied by these representations. In this way, we hope to explain more visually the correlation between visualised deep features and saliency representations. By examining the visualisation results, we aim to ascertain the extent to which the deep features may already incorporate information related to saliency and its potential impact on prediction accuracy. By visualising the pre-trained ResNet-50 and VGG-16 networks based on ImageNet~\cite{5206848}, we can explore in-depth information captured by deep features activated in different layers of the deep neural network.
\begin{figure}[ht]
    \begin{subfigure}{0.5\linewidth}
    \begin{center}
       \begin{tabular}{c} 
        \includegraphics[height=8cm]{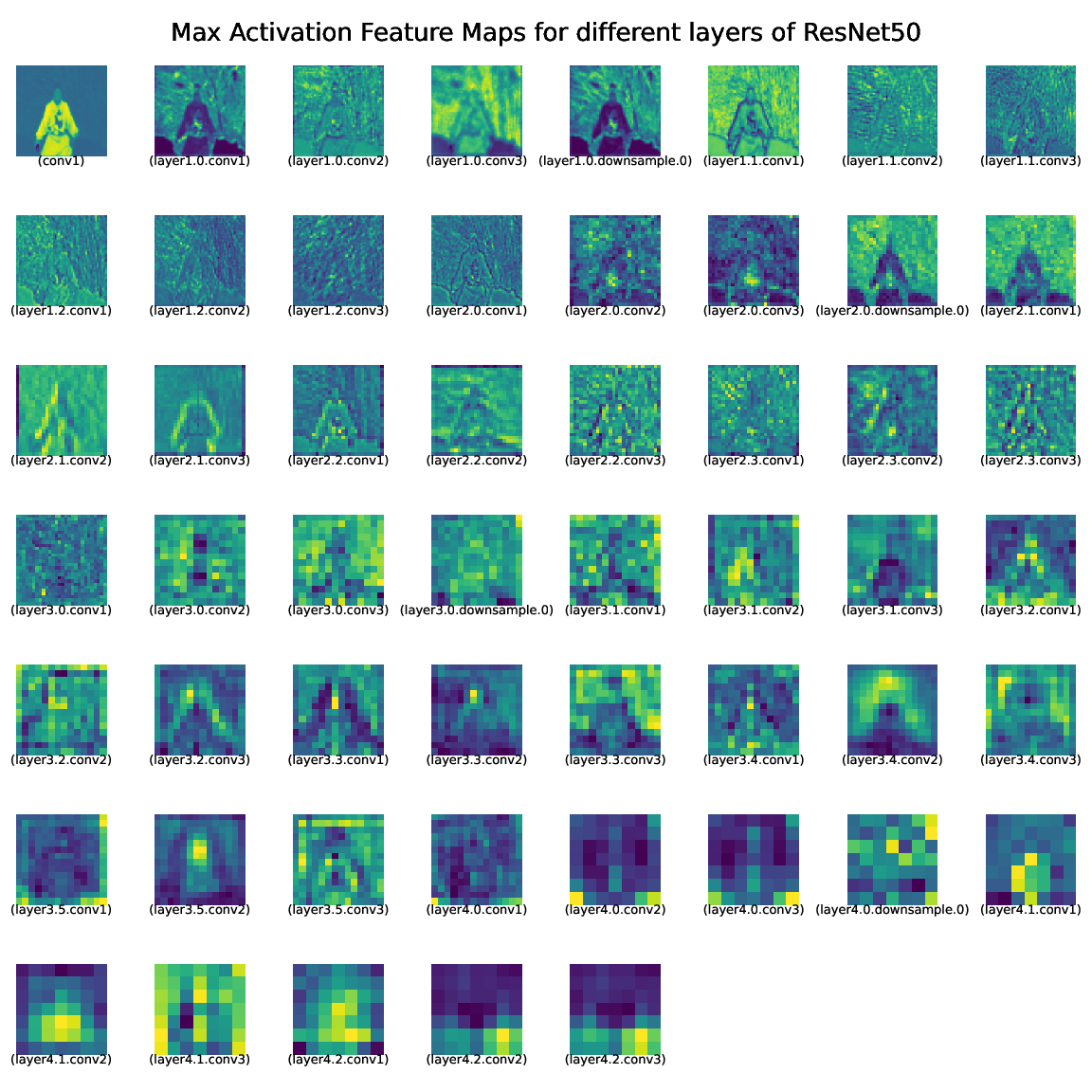}
        \end{tabular}
    \end{center}
        \captionsetup{font=footnotesize}
        \caption{ResNet-50.}
        \label{fig:resnet-max}
    \end{subfigure}%
    \begin{subfigure}{0.5\linewidth}
    \begin{center}
       \begin{tabular}{c} 
        \includegraphics[height=8cm]{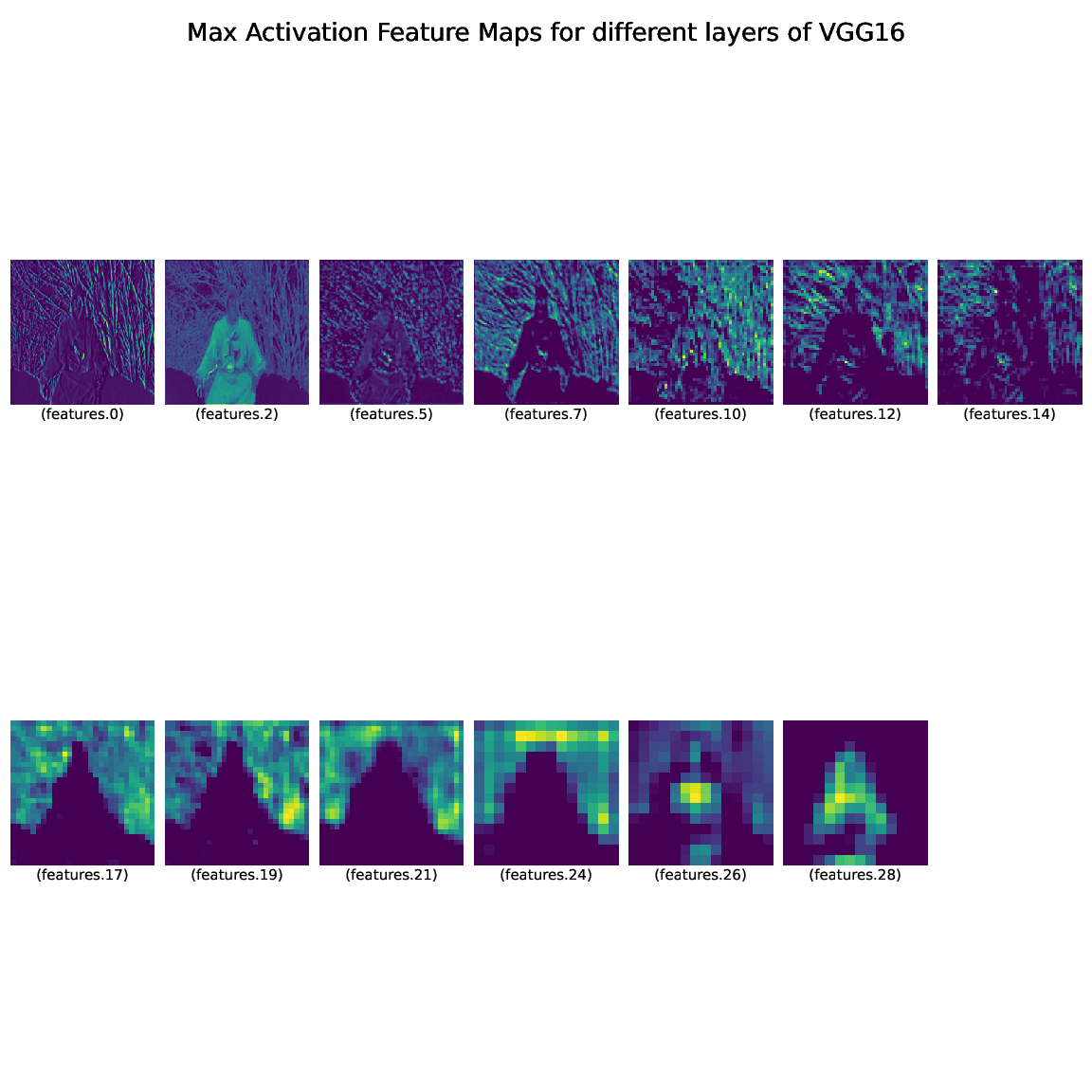}
        \end{tabular}
    \end{center}
        \captionsetup{font=footnotesize}
        \caption{VGG-16.}
        \label{fig:vgg-max}
    \end{subfigure}
    \captionsetup{font=footnotesize}
    \caption{Max activation features map across different layers for ResNet-50 and VGG-16 networks. Example of feature visualization: Sampled Frame 7 of \textbf{TelevisionClip\_1080P-68c6.mkv}, with video sampled every second (\textit{fps = 30}).}
\end{figure}
\begin{figure}[ht]
    \centering
    \begin{minipage}{0.5\linewidth}
        \centering
        \includegraphics[width=\linewidth]{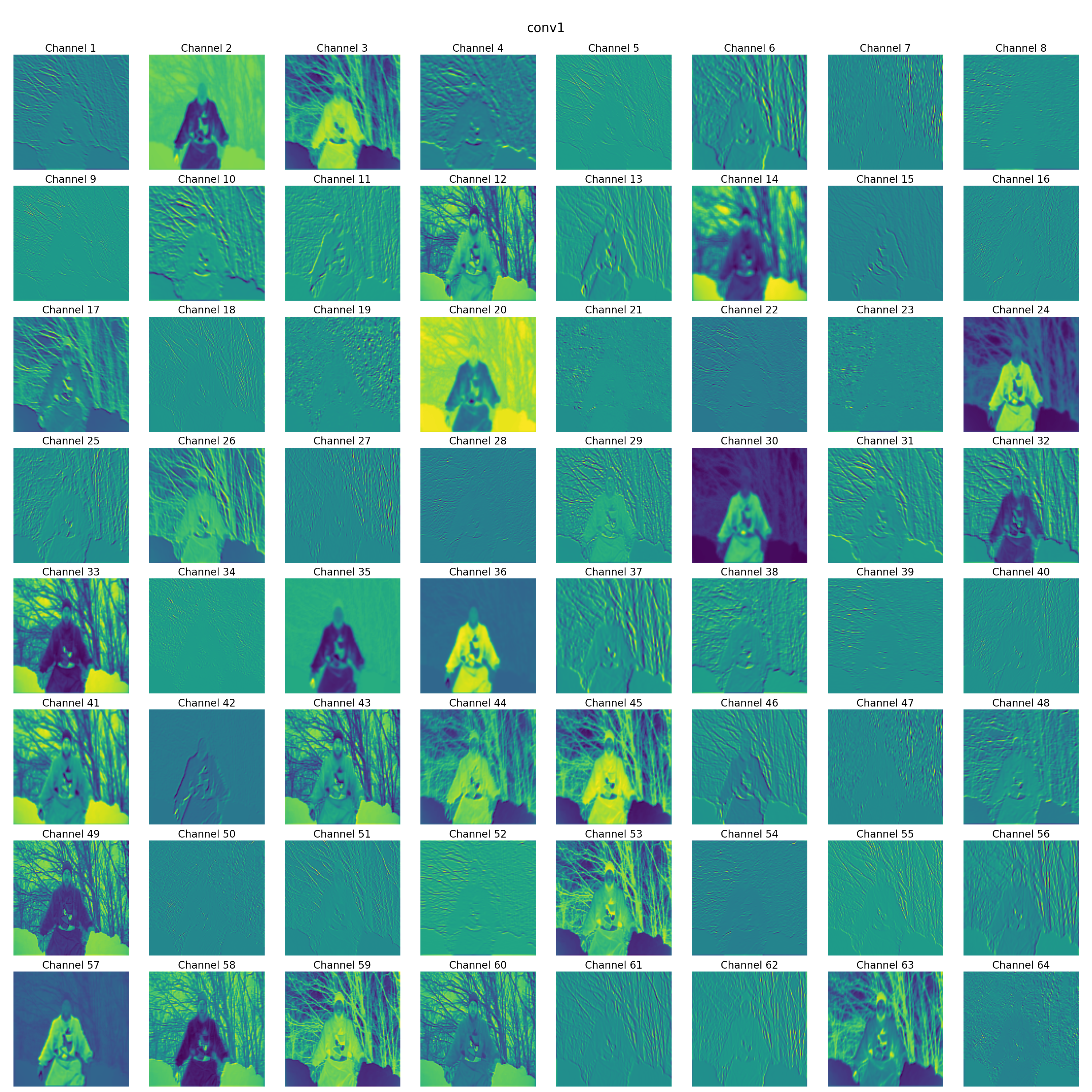}
    \end{minipage}%
    \begin{minipage}{0.5\linewidth}
        \centering
        \includegraphics[width=\linewidth]{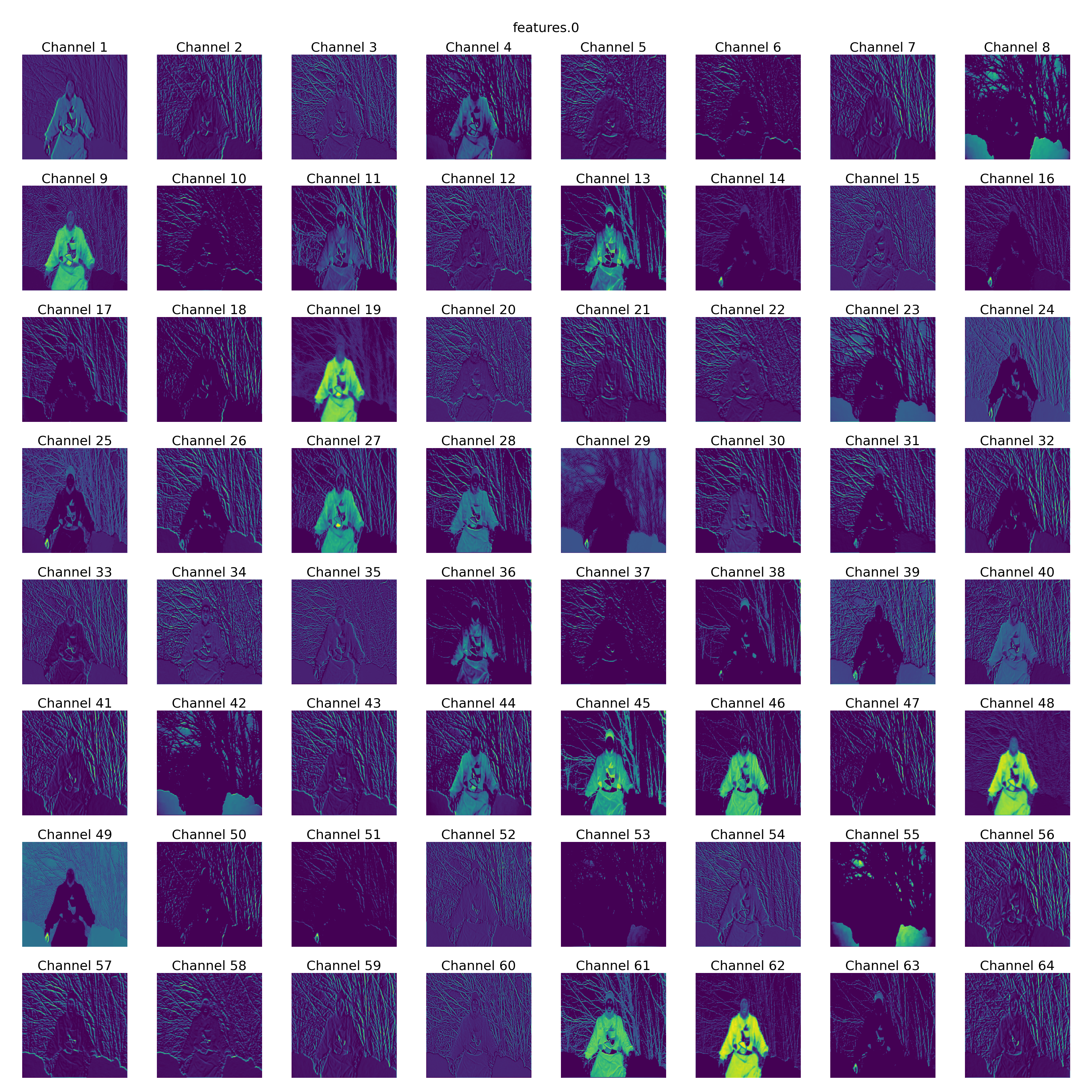}
    \end{minipage}
    \captionsetup{font=footnotesize}
    \caption[Video]{Feature maps of deep neural networks: Video 1 ResNet-50 (left) and Video 2 VGG-16 (right).}
    \label{fig:video-file}
\end{figure}

For each layer of the ResNet-50 and VGG-16 networks, we visualised and saved feature maps to gain insight into the representations learned by the networks in \textbf{Figure} \ref{fig:resnet-max} and \textbf{Figure} \ref{fig:vgg-max}. Specifically, we obtained slice images of the maximum activation channel, which allowed us to identify the most prominent features activated in each layer.
As can be seen, at the lower layers of the network, we observed low-level feature detection, such as texture, edges and holes, etc. Furthermore, we noticed that location information plays a crucial role in these layers. On the other hand, in the deeper layers of the network, we observed the extraction of more abstract features. Interestingly, location information appeared to be less critical in these deeper layers. From these activation maps, we notice that the deeper feature visualisations of ResNet-50 show higher values around similar areas picked as salient from Score-CAM in Figure~\ref{fig:resnet-max}. 

Similarly to the activation maps, by visualising and analysing deep features, we can learn how the different layers of the networks of ResNet-50 and VGG-16 encode and learn to represent visual information. In contrast to the ResNet network, we note that the VGG network retains most of the detailed information from the input image. As shown in \textbf{Figure} \ref{fig:video-file}, we provide two media files in this paper to dynamically illustrate the process of feature activation changes in different layers. These files show feature visualisations of ResNet-50 and VGG-16 networks for the same video (Sampled Frame 7 of TelevisionClip\_1080P-68c6.mkv). Due to the higher dimensionality (channels) of the deeper convolutions, here we chose to show only the first 64 channels. More information can be found in the supplementary material on the project website.

\subsection{Exploring the Effect of Saliency}
\label{subsec: effect}
By comparing the performance of metrics with and without added saliency, we started investigating whether it is likely to be implicitly represented by deep features. As a first step towards that, we pose the question: is there any correlation between the saliency maps and the DFs and how can we measure it? We first demonstrated the effect of saliency by visualising and correlating CNN deep features extracted from ResNet-50 with saliency features extracted from Score-CAM, as shown in Sec. \ref{subsec: visulisa} and after executing \textbf{Algorithms} \ref{alg:deep_feature_extraction} and \ref{alg:saliency_feature_extraction}. 
\begin{figure}[ht]
    \centering
    \begin{minipage}{0.48\textwidth}
        \begin{algorithm}[H]
            \caption{Deep Feature Extraction}\label{alg:deep_feature_extraction}
            \begin{algorithmic}[1]
                \State \textbf{Input}:
                \State An input image $I$ of size $1920 \times 1080$ down-sampled to a size $224 \times 224$
                \State \textbf{Dimension}: $(1920, 1080) \rightarrow (224, 224)$
        
                \State \textbf{Pipeline}:
                \State Image $I$ $\rightarrow$ ResNet-50 convolutional layers $\rightarrow$ Average pooling $\rightarrow$ Deep feature $DF$
               
                \State \textbf{Output}:
                \State A deep feature matrix $DF$ of size $1 \times 2048$
                \State \textbf{Dimension}: $(1, 2048)$
            \end{algorithmic}
        \end{algorithm}
    \end{minipage}
    \hfill
    \begin{minipage}{0.48\textwidth}
        \begin{algorithm}[H]
            \caption{Saliency Feature Extraction}\label{alg:saliency_feature_extraction}
            \begin{algorithmic}[2]
                \State \textbf{Input}:
                \State An input image $I$ of size $1920 \times 1080$ down-sampled to a size $224 \times 224$
                \State \textbf{Dimension}: $(1920, 1080) \rightarrow (224, 224)$
        
                \State \textbf{Pipeline}:
                \State Image $I$ $\rightarrow$  Pre-trained Score-CAM based on VGG-16 $\rightarrow$ Saliency features $S$

                \State \textbf{Output}:
                \State A saliency feature matrix $S$ of size $224 \times 224$
                \State \textbf{Dimension}: $(224, 224)$
            \end{algorithmic}
        \end{algorithm}
    \end{minipage}
\end{figure}
\begin{algorithm}
    \caption{Correlation Measurement between Deep Features and Saliency Features}
    \label{alg:correlation_measurement}
    \begin{algorithmic}[1]
        \State \textbf{Input}:
        \State Saliency feature matrix $S$ of size $224 \times 224$

        \State \textbf{Pipeline}:
        \State Dimension Reduction ($45 \times 45$) $\rightarrow$ Feature Extraction and Window Shift Calculation $\rightarrow$ Window-based Subvector Extraction($2025 \times 1$) $\rightarrow$ Equation~(\ref{eq: cosine}) Calculation $\rightarrow$ Average Correlation

        \State \textbf{Output}:
        \State Average correlation value representing the relationship between deep features and saliency features.
    \end{algorithmic}
\end{algorithm}
\begin{figure}[ht]
    \begin{minipage}{0.5\linewidth}
        \centering
        \includegraphics[height=6cm]{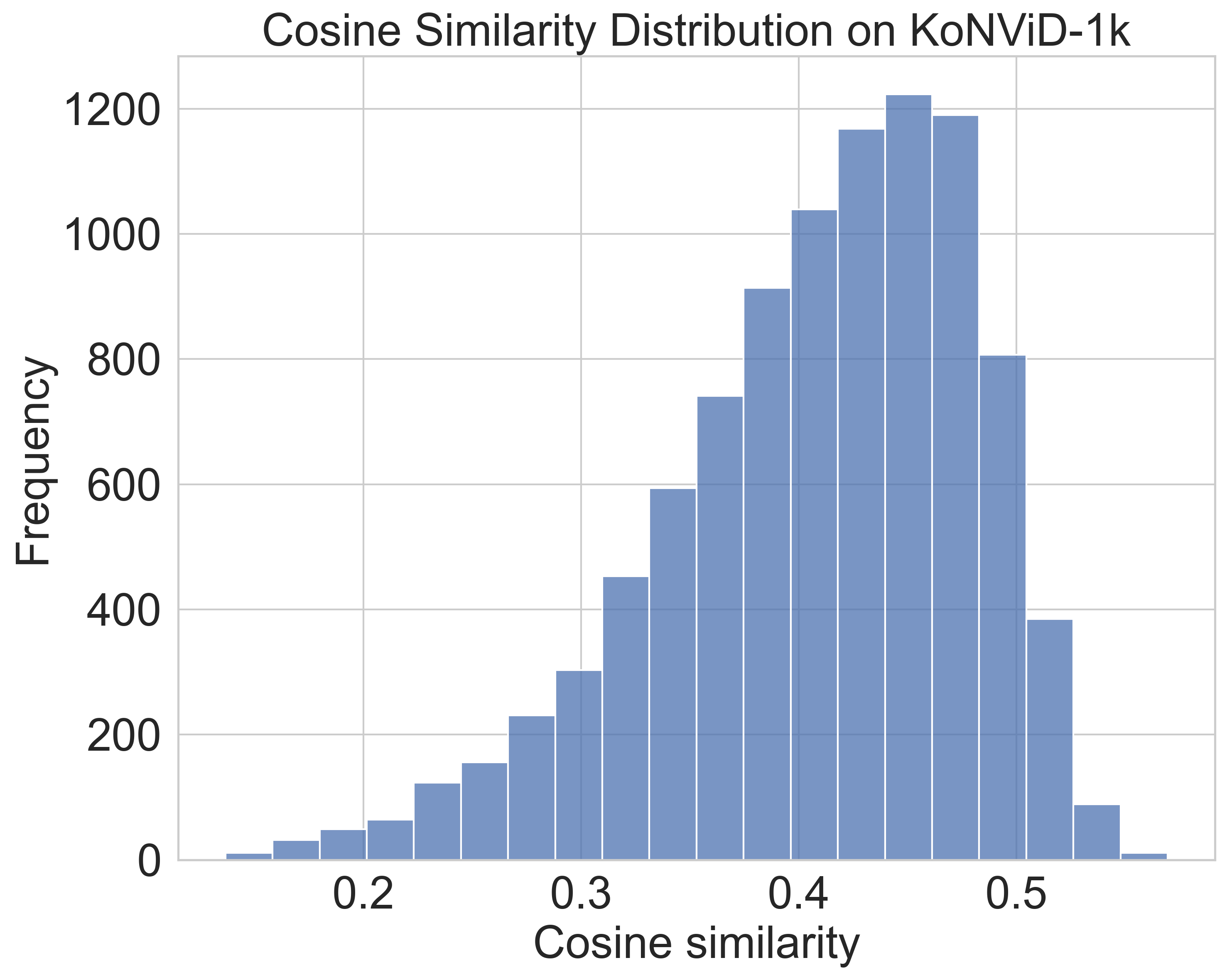}
    \end{minipage}%
    \begin{minipage}{0.5\linewidth}
        \centering
        \includegraphics[height=6cm]{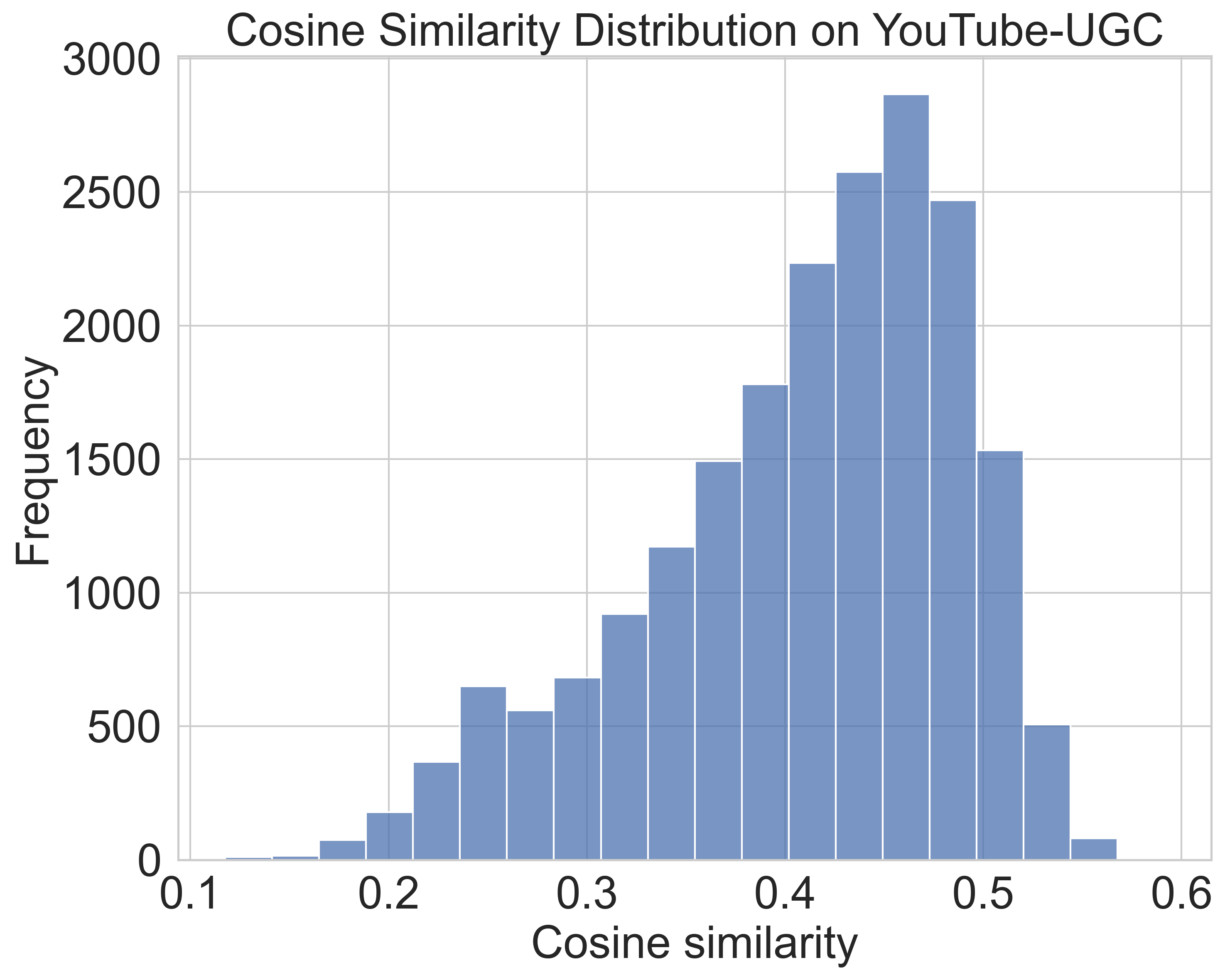}
    \end{minipage}
    \captionsetup{font=footnotesize}
    \caption{Cosine similarity between ResNet-50 DFs and Saliency maps over the two datasets.}
    \label{fig:similarity}
\end{figure}

In order to calculate the correlation of DFs, we follow the paradigm of Ayinde et al.\cite{ayinde2019correlation} who explored redundant features in deep neural network models by calculating the relative cosine distance for each layer's features in the model. Regarding the correlation value between two features, we chose to compute it by evaluating the cosine similarity between them as defined in the \textbf{Equation}~(\ref{eq: cosine}):
\begin{equation}
\label{eq: cosine}
\mathrm{Cosine}(\varphi_1, \varphi_2) = \frac{\langle \varphi_1, \varphi_2 \rangle}{\| \varphi_1 \| \|\varphi_2 \|} \; ,
\end{equation}
where $\langle \varphi_1, \varphi_2 \rangle$ is the inner product of two arbitrary normalised feature vectors $\varphi_1$ and $\varphi_2$. The $\mathrm{Cosine}(\varphi_1, \varphi_2)$ is bounded by $[-1, 1]$: The value is $1$ when $\varphi_1=\varphi_2$ or $\varphi_1$ and $\varphi_2$ are the same, $-1$ when the two vectors are in exactly opposite directions and $0$ when the two eigenvectors are orthogonal in weight space. We chose this algorithm for calculating feature similarity because cosine similarity measures the relative orientation of two vectors in the feature space, thus assessing the correlation between features. 

Because of the difference in the dimensions of DFs and saliency maps, we designed a window-shifting algorithm to enable us to measure the correlation between deep and saliency features, as shown in \textbf{Algorithm} \ref{alg:correlation_measurement}. Firstly, the dimension of the saliency feature matrix is reduced from ($224 \times 224$) to ($45 \times 45$). Subsequently, we flatten this matrix into a one-dimensional array of size ($2025,$) to serve as the window size for the window shift calculation on the deep features (size: ($2048,$)). Then, we extract the deep feature subvectors corresponding to each window and computed their cosine similarity with the corresponding saliency features. Finally, the average correlation across all windows is obtained, providing a comprehensive measure of the association between saliency features and deep features.

We extract video frames from all videos in the datasets using sparse frame sampling (one frame per second). For each video frame, we obtained two different features: deep features extracted using a pre-trained ResNet-50 model and saliency features acquired from a pre-trained Score-CAM. We calculated the cosine similarity between the two features for all video frames and analysed their distribution. 

As shown in \textbf{Figure} \ref{fig:similarity}, the distribution of the correlation between deep features and saliency is skewed, centring around the range of 0.45-0.48 across the two datasets. This demonstrates a certain degree of correlation between saliency features and DFs, indicating that they are not entirely identical or on the other hand not entirely orthogonal. 

\section{Conclusions and Future Work}
\label{sec: con}
In this paper, we explored state-of-the-art metrics for deconstructing/combining NSS and DFs. We studied whether the alignment to perceptual quality can be improved by introducing saliency maps. We used two publicly available evaluation UGC datasets to train and test the fused BVQA metrics. Preliminary results show that high accuracy can be achieved using only DFs, and adding saliency does not generally improve performance. By analysing the correlation between saliency features and deep features, we gained further insight into the impact of saliency and explored its implicit representation through DFs. More results on an expanded set of experiments can be found on our project page\footnote{\url{https://github.com/xinyiW915/SPIE-2023-Supplementary}}. 
Future research directions will focus on investigating semantic features and their significance for attention in experience quality assessment and exploring multimodal models for video quality assessment. 

\acknowledgments 
This work was funded by the UKRI MyWorld Strength in Places Programme (SIPF00006/1). This work was carried out using the computational and data storage facilities of the Advanced Computing Research Centre, University of Bristol - \url{http://www.bristol.ac.uk/acrc/}.

\bibliographystyle{spiebib} 

\end{document}